\documentclass[review]{elsarticle}

\usepackage{lineno,hyperref}
\usepackage{xcolor}
\usepackage{longtable}
\usepackage{listings}
\usepackage{hhline,colortbl,bm}
\usepackage{tikz}
\usepackage{seqsplit}
\usepackage{hyphenat}
\usetikzlibrary{trees}
\usepackage{adjustbox}
\usepackage{enumitem}
\usepackage{makecell}
\usepackage{booktabs}
\usepackage[edges]{forest}
\usetikzlibrary{shadows,arrows.meta}
\usepackage{array}
\newcommand{\PreserveBackslash}[1]{\let\temp=\\#1\let\\=\temp}
\newcolumntype{C}[1]{>{\PreserveBackslash\centering}p{#1}}
\newcolumntype{R}[1]{>{\PreserveBackslash\raggedleft}p{#1}}
\newcolumntype{L}[1]{>{\PreserveBackslash\raggedright}p{#1}}

\modulolinenumbers[5]

\begin{document}

\begin{frontmatter}

\title{A Review of Deep Learning-based Approaches for Deepfake Content Detection}

\author[1]{Leandro A. Passos\corref{c}}
\cortext[c]{Authors contributed equally.}
\ead{leandro.passos@unesp.br}

\author[1]{Danilo Jodas\corref{c}}
\ead{danilo.jodas@unesp.br}

\author[1]{Kelton A. P. Costa\corref{c}}
\ead{kelton.costa@unesp.br}

\author[1]{Luis A. Souza J\'unior}
\ead{luis.souza-junior@unesp.br}

\author[1]{Douglas Rodrigues}
\ead{d.rodrigues@unesp.br}

\author[2,3]{Javier Del Ser}
\ead{javier.delser@tecnalia.com}

\author[4]{David Camacho}
\ead{david.camacho@upm.es}

\author[1]{Jo\~ao Paulo Papa}
\ead{joao.papa@unesp.br}


\address[1]{Department of Computing, S\~ao Paulo State University\\Av. Eng. Luiz Edmundo Carrijo Coube, 14-01, Bauru, 17033-360, Brazil}

\address[2]{TECNALIA, Basque Research and Technology Alliance (BRTA), 48160 Derio, Spain}
\address[3]{Department of Communications Engineering, University of the Basque Country (UPV/EHU), 48013 Bilbao, Spain}
\address[4]{School of Computer Systems Engineering, Universidad Politécnica de Madrid, Calle de Alan Turing, 28038 Madrid, Spain}


\begin{abstract}

Recent advancements in deep learning generative models have raised concerns as they can create highly convincing counterfeit images and videos. This poses a threat to people's integrity and can lead to social instability. To address this issue, there is a pressing need to develop new computational models that can efficiently detect forged content and alert users to potential image and video manipulations. This paper presents a comprehensive review of recent studies for deepfake content detection using deep learning-based approaches. We aim to broaden the state-of-the-art research by systematically reviewing the different categories of fake content detection. Furthermore, we report the advantages and drawbacks of the examined works, and prescribe several future directions towards the issues and shortcomings still unsolved on deepfake detection.
\end{abstract}

\begin{keyword}
Fake Content\sep Machine Learning \sep Deep Learning \sep Security
\end{keyword}

\end{frontmatter}


\section{Introduction}
\label{s.introduction}

One of the major global concerns of modern society regards the development and rapid dissemination of fake information through fast-content consumption platforms, such as TikTok, Twitter, Facebook, and Instagram~\cite{kaliyar20, susan20, t20, k20}. Such content may vary from text-based messages to, most recently, image and video automatic manipulation using a family of machine learning (ML)-based approaches called deep learning.

Deep learning techniques usually stack a set of simpler ML models and apply successive operations to extract intrinsic information from data. Such approaches gathered extreme popularity in the last decade, for they achieved state-of-the-art results in virtually any field of science. Among them, deepfake content became famous in social media due to its ability to stimulate one's imagination by creating surreal and fanciful events, like presenting David Beckham speaking several languages~\footnote{Dubbing: \url{https://variety.com/2019/biz/news/ai-dubbing-david-beckham-multilingual-1203309213/} (accessed on January 31st, 2024).} (which he actually does not speak) or bringing Salvador Dal\'{i} to host visitors of Dal\'{i} Museum~\footnote{Savador Dali: \url{https://www.theverge.com/2019/5/10/18540953/salvador-dali-lives-deepfake-museum} (accessed on January 31st, 2024).}. Deepfake uses artificial intelligence to change people's faces in images and videos, synchronizing lip, eyes, and other facial expressions, as well as body movenments~\cite{liv20}. The technique is powerful enough to convey some comfort and raise nostalgic feelings by bringing some beloved people and celebrities ``back to life'', like Freddie Mercury~\footnote{Freddie Mercury: \url{https://nerdist.com/article/freddie-mercury-deepfake/} (accessed on January 31st, 2024).}. Last but not least, it also became a meme factory and source of entertainment, empowering people to ``make'' their friends and relatives to sing~\footnote{Wombo: \url{https://www.wombo.ai/} (accessed on January 31st, 2024).} and dance~\cite{chan2019everybody}, among other activities~\cite{thies2019deferred, kowalski2017deep}.

The deepfake concept spreads so fast that, nowadays, anyone equipped with a smartphone and internet access can download and manage straightforward tools to manipulate videos in any context and create realistic deepfake videos. Such a readiness raised several concerns worldwide due to the potentially negative consequences regarding unethical and malicious aspects. To cite some examples, one can find celebrities with their faces swapped with porn actrices~\footnote{Fake Porn: \url{https://www.vice.com/en/article/gydydm/gal-gadot-fake-ai-porn} (accessed on January 31st, 2024).} or tampered politicians' speeches\footnote{Obama Deep Fake: \url{https://ars.electronica.art/center/de/obama-deep-fake} (accessed on January 31st, 2024).}. In light of the evolving landscape of generative modeling choices, there is an urge to implement modern policies targeting such deepfake technologies. This need is stressed by explicitly considering such issues in emerging regulatory initiatives, such as the Artificial Intelligence Act~\cite{labuz2023regulating}, emphasizing their significant relevance in the cutting-edge technology era \cite{diaz2023connecting}.

Such a negative potential caught the attention of many researchers worldwide, proposing thousands of papers toward effective deepfake content detection using deep learning approaches. Nonetheless, a few works summarized the main challenges and technologies employed for the deepfake content detection. Nguyen et al.~\cite{nguyen2019deep} excerpted the most relevant approaches for deepfake creation and detection. Later on, Tolosana et al.~\cite{tolosana2020} provided a review on face manipulation and deepfake detection, and more recently Juefei-Xu et al.~\cite{juefei2021countering} provided a deepfake-related study exposing the battleground between deepfake generation and detection and some insights regarding tendencies and future work, while Mirsky et al.~\cite{mirsky2021creation} presented an illustrated catalog of the deepfake network architectures, also exploring the current status and tendencies of the attacker-defender game. Finally, Liz et al.~\cite{liz2024generation} compared forensic tools for generation and detection of manipulated multimodal audiovisual content using deep learning techniques. Table~\ref{t.surveys} provides a comparison among recent surveys on deepfake detection, .

\begin{table}[!htb]
\caption{Comparison of recent surveys on deepfake detection. DL stands for \emph{Deep Learning}. The number of models studied regards detection tasks only and does not consider works related to deepfake creation.}
\begin{center}
\resizebox{\textwidth}{!}{
\begin{tabular}{cccccccc}
\toprule
\textbf{Reference} & \textbf{Year}  & \makecell{\textbf{Review}\\\textbf{period}}  & \makecell{\textbf{Papers}\\\textbf{reviewed}} & \makecell{\textbf{Dataset}\\\textbf{coverage}}   & \makecell{\textbf{Models studied:}\\\textbf{Traditional/DL}} & \makecell{\textbf{Papers discussed:}\\\textbf{Traditional/DL}} & \makecell{\textbf{Future}\\\textbf{directions}}\\
\midrule
\cite{tolosana2020}& 2020 & 2016-2020 &34&Yes&Yes/Yes&9/25&Yes\\ \midrule
\cite{mirsky2021creation}& 2021 & 2017-2020&54&No&Yes/Yes&10/44&Yes\\  \midrule
\cite{juefei2021countering}& 2021 &2017-2020 &97&Yes&Yes/Yes&28/69&Yes\\  \midrule
\cite{seow2022comprehensive}& 2022 &2017-2021 &64&Yes&Yes/Yes&12/52&Yes\\ \midrule
\cite{rana2022deepfake}& 2022 & 2018-2020 &91&Yes&Yes/Yes&21/70&No\\ \midrule
\cite{nguyen2019deep}&    2022  & 2017-2021&24&No&Yes/Yes&7/17&Yes\\ \midrule
\cite{zhang2022deepfake}& 2022 & 2018-2020& 25&No &No/Yes&0/25& Yes\\  \midrule
\cite{liz2024generation} & 2024 & 2018-2023   & 66 & Yes &No/Yes&0/66& Yes\\  \midrule
Ours                    & 2024 & 2018-2024 &89&Yes&No/Yes&0/89& Yes\\ 
\bottomrule
\end{tabular}}
\label{t.surveys}
\end{center}
\end{table}



These works show that efficient deepfake detection tools are crucial, and such studies contribute to a brand new ground for research, whose demand grows to the same degree as manipulating software becomes more popular and easier to handle, leading to an increase in the number of deepfake cases and bringing several consequences to people, governments, and companies. Therefore, this survey provides an overview of the progress associated with deepfake detection techniques. It explains deepfake detection methods based on machine learning, among other intelligent systems, and also elaborates on the architectures and frameworks employed for face swapping. Further, it offers the reader a base of knowledge available in the current literature, being useful as a new source for researchers involved in image detection and security issues. Additionally, it presents a precise vision of recent research's potential challenges and the latest research guidelines. Moreover, one of the main differences and contributions is a detailed and illustrated presentation of the datasets used for detection tasks. The contributions of this work are listed below:

\begin{itemize}
    \item It provides an updated and comprehensive review of the most recent works toward deepfake creation and detection;
    \item It exposes the most recent advances, main challenges, and tendencies of the field;
    \item It presents a detailed description of the most recent and popular architectures and frameworks for deepfake creation;
    \item It supplies the reader with an illustrated catalog of datasets employed for deepfake detection.
\end{itemize}

This work is organized as follows: Section~\ref{ss.work_selection} introduces the survey methodology, search strategy, and work selection criteria employed to produce this review article. Section~\ref{s.rw} presents a collection of works on deepfake creation and detection, discussing each method's relevance and contributions, while Section~\ref{s.datasets} addresses the most recent and popular datasets used in deepfake detection research. Section~\ref{s.oi} discusses the opportunities and challenges faced by the works considered in this survey. Finally, Section~\ref{s.cn} presents conclusions and future research possibilities on deepfake detection.

\section{Review methodology}
\label{ss.work_selection}
\begin{figure}[!ht]
	\centering
	\includegraphics[width=1.0\textwidth]{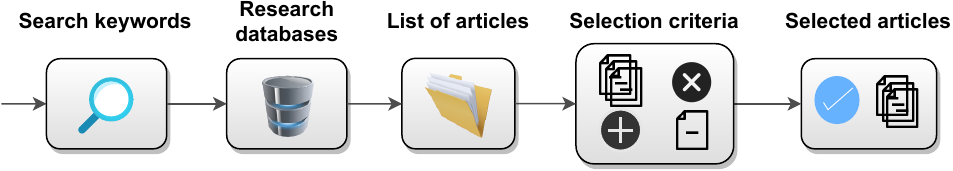}
	\caption{Proposed methodology for the literature review.}
	\label{f.review_methodology}
\end{figure}

The foremost step towards a literature review is the search for the proper studies which comprise the subject of interest. In this sense, meaningful and recent research must be selected for a complete analysis aiming at categorizing and exploring the essential works revealing the deepfake analysis and detection. The methodology employed in this review includes the steps depicted in Figure~\ref{f.review_methodology}. The following sections describe each step in detail.

\subsection{Search Keywords}
The following keywords were considered for searching the eligible articles: ``deep learning'', ``convolutional neural network'', ``deepfake detection'' , ``deepfake dataset'', ``tampering video detection'', and ``fake video detection''. Moreover, we combine the previous keywords with ``recurrent neural network'', ``LSTM'', ``GRU'', ``LLM'' and ``Transformers'' for a broader tracking of works in the temporal learning domain. A joint of keywords was used to assemble the following command search:

\begin{lstlisting}[language=TeX]
$ ("deep learning" OR "convolutional neural network" OR "LSTM" OR "GRU" OR "LLM" OR "Transformers") AND ("deepfake detection" OR "deepfake dataset" OR "tampering video detection" OR "fake video detection")	
\end{lstlisting}

\subsection{Research Databases}
The works described and analyzed in this survey were obtained through a search ranging from 2018 to the current date in the following scientific article databases: IEEExplore, ScienceDirect, ACM, Taylor $\&$ Francis Online, and Web of Sciences.

\subsection{Selection Criteria}
\label{ss.selection_criteria}
The selection criterion relied on the appraisal of the title and the abstract's content to properly examine the key features which reveal the studies focusing on deepfake creation and detection. Firstly, it considered only works published in 2018 or after since deepfake techniques gained more popularity around that period. Further, some works were left aside since they didn't fit the survey's scope in terms of application or architecture. In this context, Rezende et al.~\cite{rezende2017}, for instance, employ a shallow model, i.e., the Support Vector Machines (SVM) for classification purposes, while~\cite{shubham21,agarwal21,hoque21,Ali22,a21} presents different methods for fake image classification using distinct feature extraction techniques. Additionally, the work of Birunda et al.~\cite{selva22}, which addresses deepfake detection using the Flood Fill algorithm, can be included in this list.

In addition, coping with hoax information has also been a major concern in the context of fake news spreading on online social media in the last few years. Recent studies have been proposed to detect forgery content using natural language processing approaches and the ever-evolving deep learning architectures based on Transformers~\cite{martin2022facter} and temporal learning models~\cite{garcia2022fakenews}. Despite their outstanding importance in natural language process research, such studies were also left aside since they are beyond the scope of the topic addressed in this survey.

In summary, the works selection was based on the following inclusion standards enumerated in order of importance:

\begin{itemize}
	\item Studies conducted in 2018 or thereafter;
	\item Studies that provide public datasets containing real and forged faces;
	\item Studies comprising different methods for face manipulation on images, videos, and a joint of audio and video information;
	\item Studies applying classical machine learning algorithms;
	\item Studies applying recent deep learning models on the spatial or temporal learning domain.
\end{itemize}

The exclusion of unrelated articles was based on the following criteria:

\begin{itemize}
	\item Records related to books and conference proceedings;
	\item Systematic reviews and surveys;
	\item Studies reporting a general approach to detect any fake content in images rather than only face forgery detection;
	\item Studies reporting only the audio fake detection;
	\item Studies comprising fake news detection;
\end{itemize}

\subsection{Selected studies}
The search strategy yielded a total of 856 reports. After removing repeated records, the selected studies were eligible for the task regarding the selection criteria described in Section~\ref{ss.selection_criteria}. An initial set of 742 articles was seen as potential works in the deepfake creation and detection context. The next step involves removing the documents related to books, conference proceedings, systematic reviews, and survey articles from the document set. After examining each of the remaining 658 studies, we found 101 relevant articles meeting the standards for a full analysis and inclusion in this review. 

\section{Deepfake Detection Methods Review}
\label{s.rw}

\begin{figure}[!ht]
\centering

\begin{adjustbox}{width=\linewidth}
\begin{forest}  
forked edges,                        
for tree={grow=-90,draw,text width=2.8cm,text centered,thick,
	drop shadow,draw,rounded corners=6pt,
font=\strut\sffamily}, 
[\Large{Deepfake detection},fill=gray!40
    [\Large{Spatial Domain}\\50\%,text width=2.8cm,for tree={text centered,text width=3.8cm,fill=red!50}  
        [CNN,fill=red!35,text width=2.8cm,
            [\Large{Generative},fill=red!20,l=1cm, text centered,text width=2.8cm,
               [\textbf{Autoencoder\\}~\cite{maksutov2020,khalid2020vae,Du:2020,Venkatachalam:2022},fill=red!10,,l=2cm,text width=2.8cm]
               [\textbf{GAN\\}~\cite{hsu2019,maksutov2020,hsu20,korshunov2019vulnerability,Yang2019,frank20,umur20,Guarnera:2020,Giudice:2021,Aduwala:2021,Jeong2:2022,Varun22,Preeti:2023,Kanwal:2023,moritz22},fill=red!10,l=2cm,text width=2.8cm]
            ] 
            [\Large{Descriptive},fill=red!20,l=1cm,text width=2.8cm,
            	[\textbf{FCN}\\~\cite{sengur2018,mo2018fakeface,khodabakhsh2018,Ivanov2020,li2018exposing,afchar2018mesonet,amerini2019,gowda22,agarwal20,Agarwal2020,elrai2020,malolan20,Ranjan2020,mittal20,zi2020wilddeepfake,ciftci2020fakecatcher,wang3dcnn2021,wodajo2021,Qurat:21,joseph23},fill=red!10,text width=2.8cm]
            	[\textbf{Hybrid}\\~\cite{das2023comparative,masood2021classification,mitra2020novel,patel2020trans,rafique2021deepfake,wang2021novel},fill=red!10,text width=2.8cm]
            ]
        ]
    ]
    [\Large{Spatial+Temporal Domain}\\50\%,l=1cm,text width=5.8cm,for tree={text centered,fill=blue!20}
        [\textbf{CNN+GRU}\\~\cite{sabir2019recurrent,montserrat2020deepfakes,jaiswal2021hybrid,tu2021deepfake,hao2022deepfake,ismail2022integrated,pu2022learning,elpeltagy2023novel,sun23},fill=blue!10,l=2cm]
        [\textbf{CNN+LSTM}\\~\cite{guera2018deepfake,li18,chan2020,aldhabi2021,stanciu2021deepfake,zhang2021detecting,Ilyas2022,jalui2022synthetic,jolly2022,khedkar2022exploiting,kuang2022dual,lalitha2022,liu2022,patel2022deepfake,pipin2022deepfake,saber2022,saif2022timedist,saikia2022,saraswathi2022detection,wang2022d,Shobha23},fill=blue!10,l=2cm]
        [\textbf{Transformer\\}~\cite{khan2022hybrid,tombari22,feinland2022poker,xue22,zhang2022robust,khan2022hybrid,khormali2022dfdt,coccomini2022cross,wang2022m2tr,zhang2022deepfake,raza2023holisticdfd,heo23,lin2023deepfake,Khalid:2023,wang2023deep},fill=blue!10,l=2cm]
    ]    
]
\end{forest}
\end{adjustbox}
\caption{Taxonomy of the deepfake detection methods.}
\label{f.taxonomy}
\end{figure}
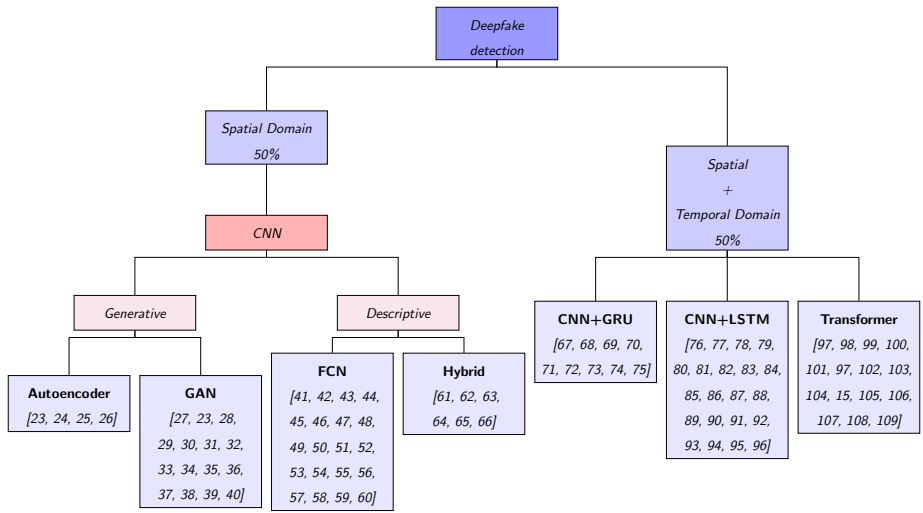

This section provides a literature review of the most recent studies containing deep learning-based approaches for deepfake detection. We categorize the works by deep learning approaches for better comprehension, i.e., Convolutional Neural Networks (CNNs) with Fully Connected Network (FCN) or hybrid approaches combining classical machine learning algorithms, Generative Models, like Autoencoder and Generative Adversarial Networks (GAN), and Recurrent Neural Networks (RNN) like Long-Short Term Memory (LSTM), Gated Recurrent Unit (GRU) and Transformer. Figure~\ref{f.taxonomy} depicts the taxonomy of the deep learning approaches reported in the literature. In summary, we can establish two main categories for deepfake detection research: spatial learning and temporal analysis.

\tikzset{tnode/.style = {basic, thin, align=left}}

Spatial learning is designed to seek evidence of facial manipulation in images and videos by using feature extraction from all video frames individually. For feature extraction, deep CNNs are usually employed to capture the feature representation of each frame in the video. Subsequently, a classification model is trained on the feature vectors for further deepfake detection at a frame-level-based approach. Eventually, the predictions are combined to determine the presence of face manipulation for the entire video. Regardless, spatial analysis usually fails to capture unnatural artifacts at the frame-dependence level along the video composition.

Unlike spatial properties, which usually capture the forgery evidence within a single image or video frames individually, temporal properties are gathered to explore inconsistencies in the video stream using the spatial features extracted from a sequence of frames, thus revealing the intercorrelation among the frames' components over time. In this sense, recurrent models have been designed to learn information dependence using a sequence of intercorrelated features, thus meeting the standards for deepfake detection via frame analysis.

\subsection{Convolutional Neural Networks}
\label{ss.cnn}

The developed works described below use similar architectures, such as Alexnet and VGG. The authors achieve significant results even with discreet architectures and different dataset changes.

Sengur et al.~\cite{sengur2018} used AlexNet and VGG16 to extract features from faces to identify fake content evidence. The proposed approach imports the trained weights via transfer learning and neglects the fine-tuning procedure, replacing the dense layer with an SVM to perform false or legit face classification. Moreover, the authors proposed combining the features obtained from both nets, providing more information and improving prediction effectiveness. The integrated features delivered the highest model performance, delivering an accuracy of $94.01\%$ on the CASIA dataset.

Meanwhile, Khodabakhsh~\cite{khodabakhsh2018} addressed the ability of some deep learning models to cope with the counterfeit face detection in images acquired from Youtube videos. The proposed study aims to appraise the model's generalizability in non-public datasets. To this extent, the authors used a dataset composed of $53,000$ images acquired from $150$ YouTube videos related to forged faces generated by CGI (Computer-Generated Imagery) and tampering methods like FakeApp and face replacement. The authors used the following popular CNN architectures: AlexNet, VGG19, ResNet, Xception, and Inception, trained on the Imagenet dataset. Despite the high accuracy obtained from Imagenet test images, the model effectiveness is severely reduced in test images of the new proposed dataset, indicating the difficulty of predicting the newly introduced artifacts. 

Gowda and Thillaiarasu~\cite{gowda22} alerted the threat of fake images and videos on various social platforms. Their work detects deep counterfeit images and videos using modified CNN models such as ResNext, Xception, and Ensemble from ResNeXt and Xception. The method achieved 80\% accuracy with ResNeXt, 78\% with Xception, and 93\% with the ensemble method.

Amerini et al.~\cite{amerini2019} presented a deepfake detection approach using optical flow vectors, calculated from two consecutive frames using a CNN-based method. The model relies on possible disorders observed in such vectors due to manipulations performed in the video. Further, the flow vectors are converted to $3$-channel color images so that VGG16 and ResNet-50 models extract features to predict a video as either real or fake.

Other approaches that use CNN models address the identification of noise, incompatibilities, and other features of the face to improve deepfake detections.

The work of Ivanov et al.~\cite{Ivanov2020} focused on the classification of counterfeit content proposing a method based on deep learning and super-resolution algorithms to expose deepfakes based on the incompatibility between the different regions of the face and the head position.

Still Li et al.~\cite{li2018exposing} developed a deep learning-based approach that aims at exposing deepfake videos by detecting face warping artifacts. The main difference from this method is that it does not require deepfake generated images as negative training examples since it targets such artifacts as the distinctive feature to detect real and fake images. The approach was evaluated over two public datasets and presented very effectively in practice. 

Agarwal et al.~\cite{agarwal20} proposed a biometric-based forensic technique for deepfake detection with static facial recognition, temporal behavior observed in facial expressions, and head movements. A CNN learns the behavioral incorporation using a metric-learning objective function. In a similar work~\cite{Agarwal2020}, the authors focused on a forensic technique for lip-sync deepfake detection. Even though the mouth shape's dynamics are sometimes inconsistent with the spoken phoneme, the method obtained state-of-the-art results. 

Mittal et al.~\cite{mittal20} presented a quantum-inspired evolutionary-based feature selection method (IQIEA-FS) to classify fake-face images. The method employs AlexNet to extract features from images and a feature selection model to discriminate the images as real and fake faces using the best features selected from the image feature vector.

Qurat et al.~\cite{Qurat:21} compared several deep CNN architectures for face forgery detection using the Real and Fake Face detection dataset~\cite{realAndFake:2019}. The work comprises image normalization and preprocessing using Error Level Analysis for further training and finetuning of different deep-learning models.

With a focus on detecting video tampering, the authors Afchar et al.~\cite{afchar2018mesonet} introduces the Mesonet, an efficient network designed to detect deepfake and Face2Face-tampered videos. The network is composed of a few layers and focuses on the mesoscopic properties of images. Experimental results show a very successful detection rate for both tasks.

Along the same lines, the work of Zi et al.~\cite{zi2020wilddeepfake} addressed deepfake detection in videos using an attention-based convolutional neural network. The model comprises 2D and 3D networks designed to use attention masks on real and manipulated faces. Moreover, the authors proposed the WildDeepfake in the very same work. In similar work, Ciftci et al.~\cite{ciftci2020fakecatcher} proposed the Fakecatcher, a deepfake detection network that employs biological signals as an implicit descriptor of authenticity. The work presented outstanding results over several public datasets, i.e., FaceForensics~\cite{rossler2018faceforensics}, FaceForensics++~\cite{rossler2019faceforensics}, and Celeb-DF~\cite{li2020celeb}, as well as a private set of data from videos in the wild.

Still Wang and Dantcheva~\cite{wang3dcnn2021} compared three distinct 3D-CNN models, namely I3D, ResNet 3D, and ResNeXt 3D,  for deepfake detection in videos. The authors considered four video manipulation techniques, providing consistent results on specific training and testing scenarios. Wodajo and Atnafu~\cite{wodajo2021} combined a CNN model with a Vision Transformer (ViT) architecture to detect videos with evidence of face manipulation. The authors rely on the VGG-16 CNN model for feature extraction from the video frames and the ViT model on such feature maps to classify the video as real or fake, obtaining significant results over the DFDC dataset.

Awotunde et al.~\cite{joseph23} addresses the considerable increase in fake videos appearing genuine thanks to advances in deepfake production tools. This investigation suggests five-layer CNNs for a DeepFake detection and classification model. ReLU-enhanced CNN extracts feature from these faces, as the model extracted the face region from the video frames. The proposed model was tested using Face2Face and DeepFake first-order motion datasets. Experimental results demonstrated that the proposed model has an average prediction rate of 98\% for DeepFake videos and 95\% for Face2Face videos in real network diffusion cases. Compared with techniques like Meso4, MesoInception4, Xception, EfficientNet-B0, and VGG16 that use CNN, the proposed model produced the most promising results with an accuracy rate of 86\%.

Mitra et al.~\cite{mitra2020novel} addressed the fake face classification in videos using a simple but effective end-to-end, fully connected deep learning architecture. The proposed method used an XceptionNet CNN as the feature extractor from the video frames. Moreover, a fully connected layer is proposed for predicting a video as authentic or fake following the fact that if one of the frames is denoted as counterfeit, the proposed method considers the entire video as deep faked. The proposed network was trained with medium compressed videos (c23 compression level) of the FaceForensics++ dataset. However, predictions on highly compressed videos showed remarkable accuracy of 93\%, while the performance on the videos with intermediate compression quality attained 96\% accuracy. However, the authors did not present the model's effectiveness on the different fake face manipulation techniques of the FaceForensics++ dataset.

Edge descriptors have also been reported as useful features for deepfake classification in videos. In this sense, Wang, Li and Zhao~\cite{wang2021novel} proposed capturing the edge information from video frames for deepfake prediction using a combined approach based on CNN for image feature extraction and the SVM as the underlying classification algorithm. Six edge filters based on four Sobel and two Laplacian operators are applied to the grayscale image of the face. Then, image feature extraction is achieved by using a ResNet-50 model in each image obtained by each edge operator. The feature maps resulting from each CNN model are concatenated and fed to a fully connected layer so that a 500-dimensional feature vector is obtained. As the final step, the SVM is used for classifying the frame as real or fake based on the 500-dimensional feature vector. The method achieved the highest AUC values against the four baseline methods used for comparison. Moreover, the authors showed the best performance attained by using the feature vector of the edge details of the frames with the SVM as the underlying classifier. In this case, the method showed an AUC of 89.3\% on the Celeb-DF dataset. However, most of the tests were performed using the Celeb-DF dataset as the underlying data for training and evaluation of the proposed approach. Though, the method also achieved a satisfactory performance on the UADFV and FaceForensics++ when the Celeb-DF dataset was used to train the models in a cross-dataset experiment.

Recent works also highlight the detection of Deep fakes by comparing frames. El Rai et al.~\cite{elrai2020} describes a deepfake detection approach through CNN and residual noise. The authors hypothesize that the residual noise obtained from the difference between an original frame and its denoised counterpart possesses strong indicators in deepfake contexts. After applying a Wavelet Transform as the denoising filter, the residual noise is computed and further used as input to an InceptionResNetV2 CNN model to detect whether the whole video is fake or not. The authors reported similar performance with two baselines in the FaceForensics dataset, thereby confirming the good effectiveness of residual noises in deepfake identification.

Furthermore Patel et al.~\cite{patel2020trans} proposed an end-to-end method combining features extracted by several CNN models for detecting deepfake videos on a frame-level-based approach. Using videos of the DFDC dataset, the authors processed the frames of the videos as individual images for further feature extraction and deepfake classification by the Random Forest classifier. The authors attained the best accuracy of 0.902 with the features extracted by the MobileNet CNN. The method is proposed for deepfake detection in a frame-level-based approach. Therefore, the temporal inter-frame correlation is not considered for the entire video classification.

Besides, a study conducted by Rafique et al.~\cite{rafique2021deepfake} addressed fake face detection in images by using two machine-learning algorithms for predicting counterfeit faces based on features extracted by AlexNet and ShuffleNet CNNs. Moreover, the authors presented a new image descriptor to strengthen the prediction capability of the proposed network. The authors assume there is a difference in compression levels of authentic and counterfeit images. In this sense, the proposed approach evaluates the difference between the original image and its counterpart version with an 85\% of compression level. The method is called Error Level Analysis (ELA), which produces an image with lossy details resulting from the compression level. The ELA image is then fed to the AlexNet and the ShuffleNet CNNs for the image feature extraction. The produced feature vector is used for the final classification as authentic or fake by SVM and $k$-NN classifiers. From experiments performed on the Real and Fake Face Detection dataset, ShuffleNet attained the highest accuracy when used as the feature extractor from the images. Moreover, combined with the $k$-NN classifier, the model provided the best-performing accuracy of 88.2\% against 87.9\% when the SVM is used as the underlying classifier.

Applicable Techniques and methods such as Transfer Learning, Generative Networks, and Fine Tuning are gaining prominence in Deepfake detection.

Malolan et al.~\cite{malolan20} explored interpretable and easily explainable models to detect deepfake videos using a deep learning-based approach. The authors trained a CNN model in a face database and applied two explainable AI techniques to visualize the image's protruding regions, i.e., the Layer-Wise Relevance Propagation (LRP) and Local Interpretable Model-Agnostic Explanations (LIME). Further, the authors presented a collection of explainable results for the model's predictions regarding heat maps, image slices, and input perturbation, indicating the model's rotational invariance and robustness to deepfake image detection. 

Ranjan et al.~\cite{Ranjan2020} analyzed three public deepfake datasets, i.e., Deepfake Detection Challenge (DFDC)~\cite{dolhansky2019deepfake}, Celeb-DF~\cite{li2020celeb}, and DeepfakeDetection (DFD)~\cite{dufour2019contributing}, which is now part of FaceForensics++~\cite{rossler2019faceforensics}, as well as a custom high-quality deepfake dataset. The work explores real-world usage scenarios through transfer learning and a deepfake detection approach based on CNNs. The authors attained 95.86\% accuracy.

Many works have applied GANs as an excellent and promising technique to detect deepfake in images and videos. In this sense, the study of Varun et al.~\cite{Varun22} explores several deepfake detection systems containing GAN with CNN to detect fake images. The authors report the latest methods to detect deep fakes made on the Internet over the years. Deep fakes are identified by training the data on two datasets. Their model achieved an accuracy of 74\% and validation accuracy of 63\% using a lightweight model.

Mo et al.~\cite{mo2018fakeface} proposed to detect fake faces using a simple CNN model based on three groups of convolutional and max-pooling layers. The authors used a set of spatial high pass filters, which perform spatial operations for highlighting fine details on images, amplifying the image's noise as a consequence. The residual noise constitutes the input features used in the proposed CNN architecture. The authors reported accuracy of $99.4\%$ in legit images of the CELEBA-HQ dataset, augmented using synthetic faces generated using a GAN-based approach~\cite{karras2017progressive}.

Other studies were also proposed towards the use of hybrid approaches combining classical machine learning algorithms with features extracted by CNN models for deepfake prediction in images and videos. Das et al.~\cite{das2023comparative} reported a frame-level-based approach for deepfake detection in videos using a hybrid strategy for feature extraction by CNN models, feature selection, and classification by a machine learning algorithm. After performing the face detection and cropping, each video frame is fed to the model for feature extraction and further classification using a classical machine learning algorithm. From each video frame used as the input image, the method combines the image features obtained by three CNN models into a feature vector that is further used for feature selection and dimensionality reduction by the Principal Component Analysis (PCA). Afterward, an SVM performs the frame classification as authentic or fake. The method attained 96.50\% accuracy on the DFDC dataset using ten components selected by PCA. It performed significantly better than the baselines end-to-end CNN models trained on the same version of the deepfake dataset. The highest score is probably due to combining feature vectors with a feature selection approach. However, the method used a reduced version of the DFDC dataset for experiments and performance evaluation. Therefore, it may not capture the variability of the deepfake traits in the full DFDC dataset.

The study of Masood et al.~\cite{masood2021classification} exploits the combination of CNN models and an SVM classifier for fake face prediction in videos. The proposed method considers a sequence of 20 video frames to perform the feature extraction and deepfake classification. The authors explored several CNN architectures in order to pick the one that performs well with the underlying classification algorithm in the feature vectors of the video frames. At last, the fake face prediction is performed by an SVM on the features extracted by the best-performing CNN model. The authors reported the highest accuracy of 98\% attained by the DenseNet-169 and the SVM classifier. Moreover, the same combination also achieved the highest values of precision, recall, and F1-Score.

Table~\ref{t.cnn} presents the summary of the methods described in this section, i.e., which considers Convolutional Neural Networks for deepfake detection. Notice that Tables~\ref{t.cnn}-\ref{t.transformers} consider the best result reported in each paper, following the best evaluation measure and the dataset whose effectiveness was the highest among the other ones.

\begin{center}
\scriptsize
\begin{longtable}{|C{.10\textwidth} | C{0.05\textwidth} | C{.13\textwidth} | C{.20\textwidth} | C{.13\textwidth} | L{.13\textwidth} |}
\caption{Summarized works considering CNN sorted by year and alphabetical order.} \label{t.cnn} \\

\hline
\multicolumn{6}{|c|}{\cellcolor{red!35}\textbf{\small Convolutional Neural Networks}} \\
\hline \multicolumn{1}{|C{.10\textwidth} |}{\textbf{Ref.}} & \multicolumn{1}{C{0.05\textwidth}|}{\textbf{Year}} & \multicolumn{1}{C{.13\textwidth}|}{\textbf{Technique}} & \multicolumn{1}{C{.20\textwidth}|}{\textbf{Dataset}} & \multicolumn{1}{C{.13\textwidth}|}{\textbf{Input}} & \multicolumn{1}{C{.13\textwidth}|}{\textbf{Best result}} \\ \hline 
\endfirsthead

\multicolumn{6}{c}%
{{\bfseries \tablename\ \thetable{} -- continued from previous page}} \\
\hline \multicolumn{1}{|p{.10\textwidth} |}{\textbf{Ref.}} & \multicolumn{1}{p{0.05\textwidth}|}{\textbf{Year}} & \multicolumn{1}{p{.13\textwidth}|}{\textbf{Technique}} & \multicolumn{1}{p{.20\textwidth}|}{\textbf{Dataset}} & \multicolumn{1}{p{.13\textwidth}|}{\textbf{Input}} & \multicolumn{1}{p{.13\textwidth}|}{\textbf{Best result}} \\ \hline 
\endhead

\hline \multicolumn{6}{|r|}{\cellcolor{gray!10}{Continued on next page}} \\ \hline
\endfoot
\hline \hline
\endlastfoot
\cite{afchar2018mesonet}  & 2018 &   CNNs & Private data  & Videos & Accuracy: 98\% \\ \hline
\cite{khodabakhsh2018} & 2018 & CNN & Fake Face in the Wild  & Videos &Accuracy: 99.60\% \\ \hline
\cite{li2018exposing} & 2018 & CNN &   UADFV, Deepfake-TIMIT& Videos & AUC: 0.999 \\ \hline
\cite{mo2018fakeface} & 2018 & CNN & CELEBA- HQ~\cite{karras2017progressive} & Fine details from high pass filters & Accuracy: 99.40\% \\ \hline
\cite{sengur2018} & 2018 & AlexNet, VGG16 & NUAA~\cite{tan2010face}, and CASIA-FASD~\cite{zhang2012face} & Images & Accuracy: 94.01\% \\ \hline
\cite{amerini2019} & 2019 & CNN & FaceForensics++ & Optical Flow from frames & Accuracy: 81.61\% \\ \hline
\cite{agarwal20} & 2020  & CNN & FaceForensics++, DFDC, Celeb-DF, WLDR~\cite{agarwal2019protecting}, and DFD~\cite{dufour2019contributing} & Videos & Accuracy: 98.90\%\\ \hline
\cite{Agarwal2020} & 2020 & CNN & Private data & Lip-sync, Audio-to-video, Text-to-video & Accuracy: 99.60\% \\ \hline
\cite{ciftci2020fakecatcher}  & 2020 &   Traditional operator+CNN &  FaceForensics, FaceForensics++, Celeb-DF, and private data & Images and Videos & Accuracy: 96\% \\ \hline 
\cite{elrai2020} & 2020 & CNN & FaceForensics and DFDC & Residual noise & Accuracy: 93.00\% \\ \hline
\cite{Ivanov2020} & 2020 & CNN + super-resolution algorithms & UADFV  & Videos & Accuracy: 95.5\% \\ \hline
\cite{malolan20} & 2020 & LRP, LIME & FaceForensics &  Faces, Images & Accuracy: 94.33\% \\ \hline
\cite{mitra2020novel}  &  2020  &   XceptionNet  &  FaceForensics++   &   Videos  &  Accuracy: 96\% \\ \hline
\cite{mittal20} & 2020 & IQIEA-FS & Real and Fake Face Detection & Images &  Mean normalized accuracy: 0.583 \\ \hline
\cite{patel2020trans}  &  2020  &   MobileNet + Random Forest  &   DFDC   &   Videos   &   Accuracy: 90.2\% \\ \hline
\cite{Ranjan2020} & 2020 & CNNs & DFD, Celeb-DF, DFDC, and private data  & Images and Videos & Accuracy: 95.86\% \\ \hline
\cite{zi2020wilddeepfake}  & 2020 &   CNNs & WildDeepfake, DFD~\cite{dufour2019contributing}, Deepfake-TIMIT, and FaceForensics++  & Images and Videos & Accuracy: 99.82\%\\ \hline
\cite{masood2021classification} & 2021  &  DenseNet-169 + SVM   &  DFDC  &  Videos   &   Accuracy: 98\% \\ \hline
\cite{Qurat:21}  & 2021 & VGG-16 & Real and Fake Face Detection & Images & Accuracry: 92.09\% \\ \hline 
\cite{rafique2021deepfake}  &   2021  &  ShuffleNet + $k$-NN  &  Real and Fake Face Detection  &  Images  &   Accuracy: 88.2\% \\ \hline
\cite{wang2021novel}  &   2021   &  ResNet-50 + SVM  &   UADFV, FaceForensics++ and Celeb-DF   &  Videos   &   AUC: 89.3\% \\ \hline
\cite{wang3dcnn2021} & 2021 &  3D-CNN  & FaceForensics++ & Videos & TCR: 87.43\% \\ \hline
\cite{wodajo2021} & 2021 & CNNs and Vision Transformers & DFDC  & Images and Videos & Accuracy: 91.5\%  \\ \hline
\cite{gowda22}  &  2022   &  CNN   &   DFDC   &   Videos   &    Accuracy 93\% \\ \hline
\cite{das2023comparative}  &  2023 & CNN + PCA + SVM  & DFDC  &  Videos  &  Accuracy: 96.50\% \\ \hline
\cite{joseph23}   &    2023   &     Five-layer CNN  &   DeepFake, Face2Face and First-Order Motion   &   Videos   &    Accuracy: 98.6\%\textsuperscript{\ddag} \\ \hline

\multicolumn{6}{l}{\textsuperscript{\ddag}Maximum score when the specified datasets are tested individually.}\\
\end{longtable}
\end{center}


\subsection{Generative Models}
\label{ss.gm}

This section covers two generative models, the Autoencoder and Generative Adversarial Network.


Maksutov et al.~\cite{maksutov2020} proposed a method to detect deepfake videos considering an artificial dataset created using GANs and autoencoders. The technique computes face features using the encoders and classifies such features using the decoders and CNNs, obtaining satisfactory values of AUC and accuracy.

Along the same lines, the work carried out by Venkatachalam et al.~\cite{Venkatachalam:2022} proposed a two-level deepfake detection in which the first phase concerns the task of extracting feature frames from the forged image using a sparse autoencoder enhanced by a graph long-short term memory and in the second phase fed the extracted features as input to a capsule network. Experiments were conducted using Flickr-Faces-HQ (FFHQ), 100K-Faces, Celeb-DF, and WildDeepfake datasets demonstrating good generalization and effectiveness in detecting deepfake images.

Khalid and Woo~\cite{khalid2020vae} proposed a Variational Autoencoder (VAE) architecture to predict fake face images in the context of one-class anomaly detection. Instead of using the binary classification task, the so-called OC-FakeDect model is trained on images of true faces that are subsequently used to predict unseen fake face images as possible anomalies. Moreover, the authors proposed a second VAE version comprising an encoder layer after the image reconstruction layer for further comparison with the latent space of the original input image's encoder. As reported in the study, the OC-FakeDect model produced better results than the binary classification task performed by the state-of-the-art Xception CNN architecture over the DFD dataset.

Du et al.~\cite{Du:2020} motivated by the degrading generalizability of deepfake detection models, they proposed a Locality-Aware AutoEncoder in which the model is forced to focus on correct forgery regions to make detection predictions. The model avoids capture dataset biases by augmenting the model with local interpretability and extra pixel-wise forgery ground truth regularization. Three types of deepfake detection tasks are proposed to evaluate the model's performance face swap manipulation, facial attributes manipulation, and inpainting-based manipulation.

Regarding Generative Adversarial Network, the work of Hsu et al.~\cite{hsu2019} combined CNN models with the contrastive loss function to cope with fake image detection. The authors combined features extracted from real and counterfeit images for the subsequent prediction by a fully connected layer attached to the feature extraction network. In the context of fake image detection, the contrastive loss may learn important aspects related to any image manipulation by comparing the features of real and forged images. Once the deep learning model is trained, it can handle the fake spots in the images' feature representation, thus achieving a high performance even in fake images generated by five GANs' architectures. The authors reported an average precision and recall of $0.88\%$ and $0.87\%$, respectively, and a maximum precision and recall of 0.947\% and 0.922\%, respectively, using the Least Squares GAN (LSGAN) for fake image generation. Later on, the same authors~\cite{hsu20} extended the work to recognize generated fake images effectively and efficiently by integrating the Siamese network with the DenseNet and contrastive loss to improve the model's performance. In this scenario, the model attained 0.968\% and 0.906\% of precision and recall, respectively, and maximum precision and recall of  0.988\% and  0.948\%, respectively, using the Progressive Growth of GANs (PGGAN) fake image generation network.

Further, Korshunov and Marcel~\cite{korshunov2019vulnerability} reported the vulnerability of state-of-the-art face recognition systems to expose deepfake videos efficiently and effectively. The authors considered several baseline approaches conducted over a custom dataset named VidTIMIT~\footnote{VidTIMIT dataset: \url{https://conradsanderson.id.au/vidtimit/} (accessed on January 31st, 2024).}. They found the best-performing methods based on visual quality metrics, often used in presentation attack detection. They also show the challenges in detecting deepfake videos produced by GANs using standard face recognition systems. Besides, they state the worst-case scenarios due to the advances of new deepfake technologies.

Besides, Yang et al.~\cite{Yang2019} developed a generative neural network-based method that splices synthesized face regions into original images, which introduces errors revealed when distinct head poses are estimated using 3D models. Such a model produces a set of features used to feed an SVM classifier for further distinguishing between real and fake images. Frank et al.~\cite{frank20} proposed a study that analyzes GAN's generated images in the frequency domain. Experimental results identified severe artifacts caused by the upsampling operations found in current GAN architectures, indicating a structural and fundamental problem in GAN's image generation procedure.

Guarnera et al.~\cite{Guarnera:2020} extracted Deepfake fingerprints from images by training an Expectation-Maximization algorithm. These fingerprints represent the Convolutional Traces left by GANs during image generation. The results demonstrated that the proposed method achieved high discriminative power and independence of image semantics considering deepfake from 10 different GAN architectures. Following the same idea, Giudice et al.~\cite{Giudice:2021} employed Discrete Cosine Transform to detect the GAN-specific frequencies and used G-boost as a classifier, demonstrating the robustness and good generalization even in deepfake videos that were not used in the training phase.

Aduwala et al.~\cite{Aduwala:2021} proposed an augmented ensemble of GAN discriminators to detect DeepFake videos. Concerning the architecture, both the GAN generator and discriminator are deep CNNs. The methodology employed consisted of a training step in which the discriminator is pre-trained, and then both the generator and the discriminator are trained together in the GAN. The experimental results demonstrated that the accuracy of the discriminator is low in unknown datasets, i.e., those datasets that did not participate in the training step. Thus, it was concluded that a GAN discriminator is not viable for handling Deepfake videos.

Jeong et al.~\cite{Jeong2:2022} designed a deepfake detector, called FrePGAN, to overcome the poor performance of GAN models on unseen data by generating frequency-level perturbation maps. These frequency-level perturbation artifacts cause the generated images to be indistinguishable from real images. Thus, at the initial iterations, the model is trained to detect these frequency-level artifacts, and at the last iterations, the model considers image-level irregularities. They employed a VGG model for the perturbation map generator, DCGAN's discriminator for the perturbation discriminator, and a pre-trained ResNet for the deepfake classifier. Experiments validated the FrePGAN as a generalized detection model reducing domain-specific artifacts in generated images.

Preeti et al.~\cite{Preeti:2023} presented a study concerning methods employed to implement deepfake. Also, they discussed deepfake manipulation and detection techniques. Furthermore, they suggested a Deep Convolution-based GAN model to detect deepfake on Deepfake Detection Challenge. The proposed model, trained in Celeb-A dataset~\cite{liu2015faceattributes}, worked well in small and limited datasets achieving higher detection capacity, i.e., telling which image is real or fake, as training iterations progress, minimizing the discriminator loss and achieving 100\% accuracy in detecting fake images.

Since there is a certain difficulty for one GAN to detect deepfake images generated by another type of GAN, Kanwal et al.~\cite{Kanwal:2023} proposed a general solution by employing siamese network with triplet loss function to detect deepfake images. Experiments were split into two cases: (i) training and test sets come from the same dataset composed of real images taken from the FFHQ dataset and fake images generated by StyleGAN; (ii) training and test sets come from different datasets in which the model is trained on FFHQ dataset and StyleGAN and evaluated on images generated by PGAN. The results prove that train using contrastive loss or triplet loss instead of cross entropy or MSE improves the generalization capacity.

In line with the previous research, Ciftci et al.~\cite{umur20} separated deep forgeries from real videos and discovered the specific generator model behind deepfake generation. The work suggests the generator's residuals contain relevant information to disentangle manipulated artifacts from biological signals. The study uses $32$ raw photoplethysmogram (PPG) signals from different face locations, encoded along with their spectral density into a spatiotemporal block, i.e., the PPG cell. The PPG cells are fed to an off-the-shelf neural network to recognize distinct signatures from the source generative models.

Complementing the studies presented, there is the research developed by Moritz~\cite{moritz22} that presents a Wavelet-packet-based analysis of GAN-generated images for deepfake analysis and detection. The authors concern the spatial-frequency properties of GAN-generated content. The method finds differences between real and synthetic images in the wavelet-packet mean and standard deviation, with rising frequency and at the edges. This mention suggests that GAN architectures must still thoroughly capture the backgrounds and high-frequency information. The same authors also found that coupling higher-order wavelets and CNN attained an improved and competitive performance compared to a Discrete Cosine Transformer (DCT) approach or working directly on the raw images, where combined architectures show the best performance.

Table~\ref{t.generative} presents the summary of the methods described in this section, i.e., using generative models to deepfake detection.

\begin{center}
\scriptsize
\begin{longtable}{|C{.10\textwidth} | C{0.05\textwidth} | C{.13\textwidth} | C{.20\textwidth} | C{.13\textwidth} | L{.13\textwidth} |}
\caption{Summarized works considering generative models sorted by year and alphabetical order.} \label{t.generative} \\

\hline
\multicolumn{6}{|c|}{\cellcolor{red!20}\textbf{\small Generative Models}} \\
\hline \multicolumn{1}{|C{.10\textwidth} |}{\textbf{Ref.}} & \multicolumn{1}{C{0.05\textwidth}|}{\textbf{Year}} & \multicolumn{1}{C{.13\textwidth}|}{\textbf{Technique}} & \multicolumn{1}{C{.20\textwidth}|}{\textbf{Dataset}} & \multicolumn{1}{C{.13\textwidth}|}{\textbf{Input}} & \multicolumn{1}{C{.13\textwidth}|}{\textbf{Best result}} \\ \hline 
\endfirsthead

\multicolumn{6}{c}%
{{\bfseries \tablename\ \thetable{} -- continued from previous page}} \\
\hline \multicolumn{1}{|p{.10\textwidth} |}{\textbf{Ref.}} & \multicolumn{1}{p{0.05\textwidth}|}{\textbf{Year}} & \multicolumn{1}{p{.13\textwidth}|}{\textbf{Technique}} & \multicolumn{1}{p{.20\textwidth}|}{\textbf{Dataset}} & \multicolumn{1}{p{.13\textwidth}|}{\textbf{Input}} & \multicolumn{1}{p{.13\textwidth}|}{\textbf{Best result}} \\ \hline 
\endhead

\hline \multicolumn{6}{|r|}{\cellcolor{gray!10}{Continued on next page}} \\ \hline
\endfoot
\hline \hline
\endlastfoot

\cite{hsu2019} & 2019 & CNN & CelebA~\cite{liu2015faceattributes} & Images & Precision: 94.70\% \\ \hline
\cite{korshunov2019vulnerability} & 2019 & VGG, Facenet & Deepfake-TIMIT & Videos & False Acceptance Rate: 95\% \\ \hline
\cite{Yang2019} & 2019 & Generative neural networks and SVM &  UADFV and DARPA MediFor GAN Image/Video Challenge~\cite{guan2019mfc} & Images and Videos & AUC: 0.89 \\ \hline
\cite{umur20} & 2020 & CNN & FaceForensics, celeb-DF, UADFV, Deepfake-TIMIT & Videos & Accuracy: 93.69\%\\ \hline
\cite{frank20} & 2020 & GANs, CNN, $k$-NN & CelebA~\cite{liu2015faceattributes} and LSUN~\cite{yu2015lsun} & Images & Accuracy: 99.91\% \\ \hline
\cite{hsu20} & 2020 & CFFN & CelebA~\cite{liu2015faceattributes} & Large pose variations, and background clutter & Precision: 98.80\% \\ \hline
\cite{khalid2020vae} & 2020 & VAE & FaceForensics++ & Images & Accuracy: 98.20\% \\ \hline
\cite{a21} & 2021 & GAN &  Private data & Images & Accuracy: 98.40\% \\ \hline
\cite{Du:2020}   &   2020   &   Locality-aware AutoEncoder  & Face Swap, Facial Attributes and Inpainting-based   &   Videos   &  Accuracy: 99.67\%\textsuperscript{\ddag} \\ \hline
\cite{Guarnera:2020}   &    2020   &  GAN fingerprint from Expectation-Minimization  &   Celeb-A   &  Videos   &  Accuracy: 93\% \\ \hline
\cite{Giudice:2021}   &   2021    &    Discrete Cosine Transform   &   Celeb-A   &  Images   &   Accuracy: 99.9\% \\ \hline
\cite{Aduwala:2021}   &   2021   &   StyleGAN-discriminator    &   DFDC, Celeb-A, 70k~\footnote{\url{https://www.kaggle.com/c/deepfake-detection-challenge/discussion/122786}} and 140k (StyleGAN)~\footnote{\url{https://www.kaggle.com/datasets/xhlulu/140k-real-and-fake-faces}}  &  Images  &  Accuracy: 92\% \\ \hline
\cite{Jeong2:2022}    &    2022   &    Frequency Perturbation GAN   &   FaceForensics++ and Custom GAN-generated~\cite{wang2020cnn}  &    Images    &     Accuracy: 79.4\% \\ \hline
\cite{Varun22}   &   2022    &   CNN+GAN   &  Celeb-A and Real and Fake Faces  &  Images  &   Accuracy: 63\% \\ \hline
\cite{Venkatachalam:2022}   &   2022   &  Sparse Autoencoder   &   FFHQ, 100K-Faces, Celeb-DF and WildDeepfake   &   Videos   &   Accuracy: 97.78\% \\ \hline
\cite{Preeti:2023}   &    2023    &    Deep Convolution-based GAN   &    Celeb-A   &    Images    &   Accuracy: 100\% \\ \hline
\cite{Kanwal:2023}    &    2023   &   Siamese Network   &     FFHQ and StyleGAN    &    Images    &   Accuracy: 94.80\% \\ \hline
\cite{moritz22}   &  2023    &   Wavelet-packet   &   FFHQ, Celeb-A, Large-scale Scene UNderstanding (LSUN) and FaceForensics++   &   Images   &    Accuracy: 96.91\% \\ \hline

\multicolumn{6}{l}{\textsuperscript{\ddag}Maximum score when the specified datasets are tested individually.}\\
\end{longtable}
\end{center}

\subsection{Recurrent Neural Networks}
\label{ss.RNN}

\begin{figure}[!ht]
	\includegraphics[width=.96\textwidth]{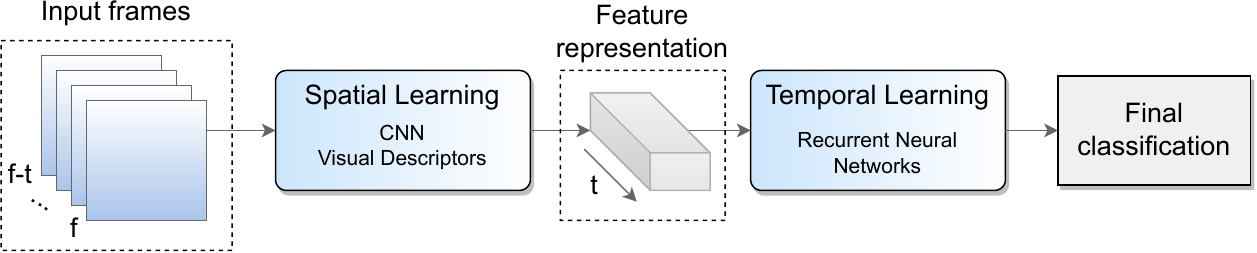}
	\caption{General strategy for the learning of temporal sequences of $t$ frames in videos.}
	\label{f.rnn_illustration}
\end{figure}

Deep learning models applied to spatial properties of the images and videos are usually unable to effectively capture the artifacts changes and inter-correlation among the frames sequence of the video. One strategy regards classifying each video frame individually and taking the most common class for the whole video classification as real or a forgery by manipulation. However, this approach may not find the connection among the aspects that lead to a deepfake generation in high-quality and realistic deepfake videos. In contrast to spatial learning performed by classical deep learning models, temporal learning provides a reasonable strategy for capturing the intrinsic aspects that compose the traits of face manipulation across a series of visual information. In this sense, temporal learning models like recurrent neural networks can handle the drawbacks of a single-frame classification and reach a consensus on the entire video classification. In this approach, each video frame is fed to a recurrent model for learning the dependency among the visual traits of the face in a sequence. Afterward, the outputted temporal representation is handed by a model for the final classification of the whole video. By doing so, we can achieve a more effective forgey identification than only using the spatial features from the video frames. Figure~\ref{f.rnn_illustration} depicts the general process of the temporal learning approach for deepfake classification in videos.

Long-Short Term Memory (LSTM) and Gated Recurrent Unit (GRU) are the two widespread structures for temporal learning in sequences of data. LSTM is a special recurrent neural network model proposed to cope with the gradient vanishing in long-term dependency problems. The LSTM comprises three elements called gates: an input gate, a forget gate, and an output gate. The input gate receives the information and updates the current state of the model. The forget gate determines the irrelevant information that should be discarded, while the output gate provides the updated information. By doing so, LSTM can handle long-term sequences by deciding what information should be retained or updated. As a recurrent neural network-inspired model, GRU can also cope with long data series by combining resetting and updating mechanisms. However, unlike the LSTM structure, the GRU architecture incorporates only the update and reset gates. The update gate determines the portion of the past information that must be passed through the next layers. On the other hand, the reset gate is responsible for deciding the information that should be neglected. In this fashion, GRU reveals less number of parameters than LSTM.

Several studies have reported LSTM and GRU-based approaches as an alternative strategy to cope with the deepfake classification. The following sections describe the most recent studies proposed for classifying deepfake videos by combining spatial features and LSTM and GRU mechanisms for temporal learning.

\subsubsection{LSTM}
G{\"u}era and Delp~\cite{guera2018deepfake} proposed a temporal-aware system for automatically detecting deepfake videos using CNNs for frame-level feature extraction and further feed a recurrent neural network for classification. The method is evaluated considering a large set of deepfake videos obtained from various websites, achieving competitive results for the task.

Similarly, Li et al.~\cite{li18} described a method that exposes videos with fake faces generated from deep neural network models. This method detects the blink of an eye in videos, usually not treated in fake videos. The process combines Convolutional Neural Networks and Long-term Recurrent CNNs (LRCN) to distinguish between open and closed eye states. 

False information provided through algorithmically modified footage, images, audio, and the emergence of misinformation from fabricated content requires the development of anti-disinformation methods such as deepfake detection algorithms to verify the validity of digital content. To cope with such a task, Chan et al. \cite{chan2020} proposed a blockchain Hyperledger Fabric 2.0 designed with LSTMs for audio/video/descriptive captioning to combat deepfake media. The framework combines various LSTM networks to trace and track digital content’s historical provenance. As an outcome and contribution to cope with deepfake scope, discriminative features created by a deep encoder allow proof of authenticity (PoA) for digital media using a decentralized blockchain of multiple LSTMs.

Considering the progressive quality of deepfake information created by deep learning techniques, better algorithms to detect them are highly demanded. Al-Dhabi and Zhang \cite{aldhabi2021} then presented a solution based on a combination of CNN and RNN, whose research highlights that using a CNN and RNN combined architecture achieves promising results. From a pre-trained ResNeXt50, the time of the model's training is saved from scratch and used for feature extraction from the video frames. The feature maps are then used to train the LSTM blockchain. As the authors concluded, the CNN and RNN combination captures the inter and intra-frame features to detect if a video is real or fake. Using a large collection of deepfake videos gathered from various distribution sources, the authors demonstrated their model's performance with around 95.5\% accuracy in the positive deepfake detection, proposing competitive results when employing a simplistic architecture.

From deep generative algorithms, such as GAN, Saikia et al.\cite{saikia2022} proposed an approach to synthesize pseudo-realistic videos that usually are very difficult to distinguish. In most cases, CNN based discriminators are used to detect such synthesized media. However, it primarily emphasizes the spatial attributes of individual video frames, thereby failing to learn the temporal information from their inter-frame connections. To cope with the hard task of learning temporal information from video's inter-frame relations,  the authors employed an optical flow-based approach to extract temporal features, then fed to a hybrid model for classification composed of CNN and RNN architecture combination. Such a hybrid model showed effective performance on the tested open source datasets, such as DFDC, FF++, and Celeb-DF, with an accuracy of 66.26\%, 91.21\%, and 79.49\%, respectively,  with a very reduced amount of samples (less than 100 frames), outperforming other works for the same fake detection modality.

To propose a new way of detecting deepfake in continuous frame video samples, Liu et al. \cite{liu2022} presented a robust deepfake video detection method, namely EfficientNet-LSTM, based on steady frame face-swapping. From an in-house face-swapping dataset with Delaunay triangulation and piecewise affine transform (to ensure the continuous face-swapping fashion), the authors described facial and background information using (i) EfficientNet to extract intra-frame fusion features and (ii) LSTM to extract inter-frame time features, composing a final mask fusion zone based on both. The cross-domain experiments highlight that the proposed method outperforms previous ones, with higher AUC values of 84.38\%.

To introduce a fully-automatic and efficient approach to getting facial expressions in videos, Jolly et al. \cite{jolly2022} proposed a model aiming to detect deepfake or synthetic information from recorded frames. Employing the FaceForensics++ dataset for training the model composed of a Residual-Net as the backbone (feature extraction) and an LSTM module to build a temporal sequence for face manipulation between frames, the authors achieved more than 99\% successful detection rate in Deepfake, Face2Face, FaceSwap, and neural texture. Thus, the approach proposed by the authors is composed of (i) detecting the subject's existing facial region by extracting and processing features using a CNN and LSTM combined model. Meanwhile, the Recycle-GAN was employed to merge spatial and temporal data. 

Lalitha and Kavitha's work \cite{lalitha2022} proposed a robust neural network-based method to identify deepfake videos. A model with a main goal of detecting artifacts and composed of a CNN and a classifier layer based on GAN technology is designed,  followed by a head of a Resnet, ResNeXt50, or LSTM  in favor to decide which structure to pair with the classifier while detecting the fake frames. The subsequent classifier network uses CNN's feature vectors to categorize whether a video is fake or real. The dataset considered comes from DeepFake Detection Challenge. Compared to previous state-of-the-art studies, the key video frame extraction method decreases computations by achieving 97.2\% accuracy on the DeepFake Detection Challenge dataset.

To determine the rightfulness of a video, Saber et al. \cite{saber2022} used and compared several deepfake detection techniques to detect fake videos. By applying techniques such as YOLO-CRNN, LSTM, etc., the authors compared the models' performances by employing EfficientNet-B5 to extract spatial features from faces on video recordings, feeding them as a batch of input series into a two-way long- and short-term memory (BiLSTM). The proposed assessment is then tested on CelebDFFaceForencics++ (c23), a dataset based on a mash-up of two well-known records: FaceForencics++ (c23) and Celeb-DF. As a result, the authors achieved AUC outcomes of 89.35\%, an accuracy of 89.38\%, a recovery of 83.13\%, and an F1-measure of 84.23\% to insert data focus.

Patel et al.~\cite{patel2022deepfake} proposed a joint spatial and temporal learning approach for deepfake detection in videos by combining a CNN model and an LSTM network to further detect the face as authentic or manipulated by generative models. In their study, the authors proposed a new joint dataset comprising 50\% real and 50\% fake videos collected from YouTube and the FaceForensics++ dataset. The method considers up to 100 video frames for further feature extraction and temporal analysis by an LSTM layer. At last, a detection network predicts if the video is authentic or manipulated. The highest accuracy (91.50\%) was attained by using 80 sequences of video frames. Moreover, a web interface has been designed to upload the video for the subsequent deepfake prediction.

Kuang et al.~\cite{kuang2022dual} explored a dual-branch approach to capture inconsistencies from the sequence of video frames to detect deepfake manipulation in videos. The method comprises spatial and temporal branches for learning the spatial and temporal information from the input video. The authors used the EfficientNet-B0 to capture the feature maps from the sequence of the video frames in the spatial branch. Then, it is followed by a fully connected layer that predicts if the video is genuine or fake. At last, the method takes the average scores of all the video frames to compute the final prediction for the video. The proposed model attained a maximum accuracy of 98.21\% in detecting forgery faces produced by deepfake methods in the FaceForensics++ dataset. Moreover, experiments conducted on the Celeb-DF dataset also showed that the proposed model attained the best performance against the baseline models used for comparison, providing an accuracy of 98.93\%. 

In a similar study, Wang, Li and Zhao~\cite{wang2022d} proposed a dual-stream and a dual-utilization network that is firstly pre-trained on real and deepfake videos for the subsequent frame feature extraction and deepfake classification by an SVM classifier. In their work, dual-stream refers to the joint spatial learning and temporal learning used in combination for feature extraction from a sequence of video frames. A spatial branch was proposed to learn the edge information obtained by six edge operators applied to the video frames. Then, a binary classification is performed by a fully connected network to predict the video as authentic or manipulated. Dual-utilization is the process of training and learning the intrinsic features from the video frames in the dual-stream domain, followed by the subsequent feature extraction for further classification by the SVM classifier. The method achieved the highest accuracy of 96.2\% against several baseline deep learning models adopted for comparison in deepfake detection. 

Shobha et al.~\cite{Shobha23} proposed using spatial learning and temporal analysis for deepfake detection by training the deep learning models on a large dataset of real and counterfeit videos. A comparative study of different models was evaluated using the Celeb-DF and Face Forensic++ datasets. The web-based framework using Python is designed to upload a video and detect deep fakes by implementing deep neural networks. The proposed method categorizes videos more precisely to establish whether a video is real or fake. By combining ResNet-50 CNN and LSTM layers, the proposed method can help leverage the strengths of both architectures and enhance the accuracy of deepfake detection by involving both image-based and sequential data. The proposed approach attained a maximum accuracy of 87.48\% in 40 epochs for the model's training.

Pipin et al.~\cite{pipin2022deepfake} addressed the deepfake detection in videos by combining a Deep Learning algorithm and Photo-Response Non-Uniformity pattern for the noise analysis of the video frames. Deep learning is modeled using ResNeXt-50 and LSTM and Photo-Response Non-Uniformity analysis (PRNU) to check the PRNU pattern of each frame in a video for deep fake prediction with high accuracy value reaching 97.89\% using 100 input video frames of the FaceForensics++ dataset.

The paper of Stanciu et al.~\cite{stanciu2021deepfake} proposed using a spatiotemporal CNN-LSTM approach for deepfake detection in videos using three selected facial regions. The study compares the model's performance in combined facial areas like the nose, mouth, and eyes and the entire face on two datasets. The proposed approach shows significant improvements when using a temporal network provided with 60 video frames as the input sequence for the deepfake detection; the method yields a 13.46\% increase in AUC for the Celeb-DF dataset (from 83.6\% to 97.06\%) and a 99.95\% AUC for the FaceForensics++ dataset.

Ilyas et al.~\cite{Ilyas2022} stated the challenges in detecting deepfake videos because of the temporal features that might differ between the video frames. In addition, frame-level visuals are becoming more realistic due to a tiny imperceptible modification in each frame. Due to these aspects, the authors introduce a hybrid deep learning model called InceptionResNet-BiLSTM, which employs the customized InceptionResNetV2 as a front-end feature extractor and Bidirectional Long-Short Term Memory (BiLSTM) network as a back-end classifier. The model extracts the features from the frames of the videos by employing a customized InceptionResNetV2 and then passes the feature vectors to the temporally aware Bidirectional LSTM, which simulates the class dependency in forward and backward directions. The authors attained an accuracy of 93.39\% in videos manipulated by deepfake techniques in the FaceForensics++ dataset.

Zhang et al.~\cite{zhang2021detecting} presented a temporal learning strategy for coping with deepfake detection in videos by using facial features and an LSTM network. The authors proposed the Facial Alignment LSTM (FA-LSTM) and the Dense Face Alignment LSTM (DFA-LSTM) to extract facial features from videos for the subsequent classification as real or fake. The facial traits are based on 68 landmark points obtained from the facial alignment method and 3D dense features extracted by the DFA method. Each facial component is independently used to train a bidirectional LSTM model for the temporal learning of the interconnection between the video frames. The authors reported 0.932 and 0.941 accuracies for the FA-LSTM and DFA-LSTM, respectively, on videos of high-quality compression of the FaceForensic and FaceForensics++ datasets. 

Jalui et al.~\cite{jalui2022synthetic} proposed a method for deepfake detection in videos by using the ResNeXt-50 CNN and an LSTM layer for the spatial learning and the temporal analysis of the frame's feature vectors. After extracting the features from the video frames, an LSTM layer receives the 2,048-dimensional feature vectors to learn the visual sequences and further classify the video as authentic or fake. The authors used only 550 videos from the Deepfake Detection Challenge (DFDC) dataset to train and validate the proposed deepfake detection approach. The model achieved 96.36\% accuracy on 110 samples of the test set. A correlation-based strategy was also employed to neglect similar frames from the videos. However, the number of frames used as input by the LSTM model was not detailed in the study.

In a similar study conducted by Saraswathi et al.~\cite{saraswathi2022detection}, a deepfake detection method was proposed by combining spatial learning and temporal analysis with a CNN and an LSTM network. In the proposed approach, a pre-trained ResNeXt-50 CNN was also used for the feature extraction from each video frame. The authors used a sequence of 20 video frames for feature extraction and deepfake classification by an LSTM model. The feature vector received by the LSTM layer is 2,048 in dimensional size. Different from the study of Jalui et al.~\cite{jalui2022synthetic}, the authors used a mixture of videos from Celeb-DF, FaceForensics++, and Deep Fake Detection Challenge (DFDC) datasets for training and validating the proposed approach. The proposed method attained 90.37\% accuracy on the test set of the created dataset.

Khedkar et al.~\cite{khedkar2022exploiting} proposed a CNN-LSTM architecture for the deepfake classification in videos. The authors used four pre-trained CNN models, namely VGG-19, ResNet-50 v2, Inception v3, and DenseNet-121, for feature extraction from 40 video frames before the temporal learning by two LSTM layers. Then, a dense layer is used for the final classification. The method was tested in the Face Forensic++ and DFDC datasets. The proposed model yielded 0.908 AUC and 90.7\% accuracy with the frame's spatial representation obtained by the DenseNet-121 and two LSTM layers for temporal learning.

Saif et al.~\cite{saif2022timedist} proposed a method for face forgery detection in videos by a deep temporal learning architecture based on LSTM. The authors used the contrastive loss function for the cross-learning aspects of pairs of real and faked video frames. Moreover, several CNN architectures were tested for feature extraction from the video frames. EfficientNet B3 attained the highest accuracy when compared to the other CNN backbones for feature extraction, providing 97.3\% accuracy on videos forged by deepfake techniques and 91.36\%, 91.85\%, and 88.15\% for FaceSwap, Face2Face, and NeuralTexture manipulation, respectively. However, it performed less than most baseline models on the entire FaceForensics++ dataset. Nonetheless, the model attained the best performance with 90.95\% and 98.7\% accuracy on videos with low and high-quality compression, respectively. 

Table~\ref{t.lstm} presents the summary of the methods described in this section, i.e., using LSTM to deepfake detection.

\begin{center}
\scriptsize
\begin{longtable}{|C{.10\textwidth} | C{0.05\textwidth} | C{.13\textwidth} | C{.20\textwidth} | C{.13\textwidth} | L{.13\textwidth} |}
\caption{Summarized works considering LSTM sorted by year and alphabetical order.} \label{t.lstm} \\

	\hline
\multicolumn{6}{|c|}{\cellcolor{blue!10}\textbf{\small CNN+LSTM}} \\
	\hline \multicolumn{1}{|C{.10\textwidth} |}{\textbf{Ref.}} & \multicolumn{1}{C{0.05\textwidth}|}{\textbf{Year}} & \multicolumn{1}{C{.13\textwidth}|}{\textbf{Technique}} & \multicolumn{1}{C{.20\textwidth}|}{\textbf{Dataset}} & \multicolumn{1}{C{.13\textwidth}|}{\textbf{Input}} & \multicolumn{1}{C{.13\textwidth}|}{\textbf{Best result}} \\ \hline 
	\endfirsthead

\multicolumn{6}{c}%
{{\bfseries \tablename\ \thetable{} -- continued from previous page}} \\
\hline \multicolumn{1}{|p{.10\textwidth} |}{\textbf{Ref.}} & \multicolumn{1}{p{0.05\textwidth}|}{\textbf{Year}} & \multicolumn{1}{p{.13\textwidth}|}{\textbf{Technique}} & \multicolumn{1}{p{.20\textwidth}|}{\textbf{Dataset}} & \multicolumn{1}{p{.13\textwidth}|}{\textbf{Input}} & \multicolumn{1}{p{.13\textwidth}|}{\textbf{Best result}} \\ \hline 
\endhead

\hline \multicolumn{6}{|r|}{\cellcolor{gray!10}{Continued on next page}} \\ \hline
\endfoot
\hline \hline
\endlastfoot
\cite{guera2018deepfake} & 2018 & InceptionV3, LSTM & Videos from multiple websites & Videos & Accuracy: 94.00\% \\ \hline
\cite{li18} & 2018 & CNN and EAR & CEW~\cite{song2014eyes} and EBV~\cite{li18} & Videos & Accuracy: 99.00\% \\ \hline
\cite{chan2020}  &  2020   &   Blockchain Hyperledger Fabric  &   N/A   &   Videos   & N/A \\ \hline
\cite{aldhabi2021}  &   2021   &   ResNeXt50   &   DFDC, FaceForensics++ and Celeb-DF   &   Videos   &   Accuracy: 95.5\% \\ \hline
\cite{stanciu2021deepfake}   &   2021   &    Xception    &    Celeb-DF and FaceForensics++   &    Videos    &   AUC: 99.95\%\textsuperscript{\ddag} \\ \hline
\cite{zhang2021detecting}   &   2021   &   Dense Face Alignment   &    FaceForensics and FaceForensics++   &   Videos    &    94.10\%\textsuperscript{\ddag} \\ \hline
\cite{Ilyas2022}    &    2022    &  Inception ResNetV2   &   FaceForensics++   &   Videos   &   Accuracy: 93.39\% \\ \hline
\cite{jalui2022synthetic}   &   2022    &   ResNeXt-50   &   DFDC   &   Videos    &     Accuracy: 96.36\% \\ \hline
\cite{jolly2022}  &   2022   &   ResNet18   &  FaceForensics++ &  Videos  &  Accuracy: 99.26\% \\ \hline
\cite{khedkar2022exploiting}   &   2022   &    DenseNet-121   &    DFDC and FaceForensics++   &   Videos   &   Accuracy: 90.7\% \\ \hline
\cite{kuang2022dual}   &   2022   &    EfficientNet-B0 + SVM  &  FaceForensics++ and Celeb-DF &  Videos   &   Accuracy: 98.93\%\textsuperscript{\ddag} \\ \hline
\cite{lalitha2022}  &   2022   &   ResNeXt50   &   FaceForensics++ and DFDC    &   Videos   &   Accuracy: 97.2\% \\ \hline
\cite{liu2022}  &  2022   &   Delaunay traingulation + Piecewise affine + EfficientNet &  FaceForensics++ and Celeb-DF   &   Videos   &   AUC: 84.38\% \\ \hline
\cite{patel2022deepfake}   &   2022   &    ResNeXt   &   FaceForensics++ and YouTube videos\textsuperscript{\dag}   &   Videos   &   Accuracy: 91.50\% \\ \hline
\cite{pipin2022deepfake}  &   2022    &   ResNeXt-50 + PRNU  &   FaceForensics++  &  Videos   &   Accuracy: 97.89\% \\ \hline
\cite{saber2022}   &   2022   &   EfficientNet-B5   &   FaceForensics++ and Celeb-DF\textsuperscript{\dag}   &    Videos   &  Accuracy: 89.38\% \\ \hline
\cite{saif2022timedist}   &   2022   &   EfficientNet-B3   &   FaceForensics++   &   Videos    &    Accuracy: 97.3\% \\ \hline
\cite{saikia2022}  &   2022   &   Optical flow + VGG-16   &  DFDC, FaceForensics++ and Celeb-DF   &   Videos   &  Accuracy: 91.21\%\textsuperscript{\ddag} \\ \hline
\cite{saraswathi2022detection}  &   2022   &   ResNeXt-50  &   Celeb-DF, DFDC and FaceForensics++\textsuperscript{\dag}  &  Videos   &   Accuracy: 90.37\% \\ \hline
\cite{wang2022d}   &   2022   &  Edge descriptors + ResNet + SVM  &  FaceForenscis++ and Celeb-DF\textsuperscript{\dag}  &  Videos  &   Accuracy: 96.2\% \\ \hline
\cite{Shobha23}   &   2023   &  ResNet-50  &   Celeb-DF and FaceForensics++   &  Videos   &   Accuracy: 87.48\% \\ \hline 

\multicolumn{6}{l}{\textsuperscript{\dag}Experiments conducted over a mash up of the specified datasets.} \\
\multicolumn{6}{l}{\textsuperscript{\ddag}Maximum score when the specified datasets are tested individually.}\\
\end{longtable}
\end{center}

\subsubsection{GRU}
In the work of Sabir et al.~\cite{sabir2019recurrent}, a recurrent neural network approach was proposed to address the deepfake detection using CNN and GRU layers for feature extraction and temporal learning of the video frames, respectively. Moreover, the authors addressed the face alignment among the video frames through a landmark-based alignment method and a spatial transformer network to learn the spatial parameters for the affine transformation and face alignment. The use of DenseNet CNN with face alignment and a GRU layer attained significant improvements and the best prediction accuracy when Face2Face and FaceSwap manipulations were applied to the videos of the FaceForensics++ dataset. In contrast, the deepfake manipulation detection is slightly better when the three components are combined together, achieving 96.9\% accuracy compared to the 96.7\% accuracy provided by the DenseNet and the face alignment method without the GRU layer. It shows the difficulty in predicting high-quality and sophisticated manipulation when deepfake methods are applied to the videos. Moreover, the landmark-based alignment strategy attained the best accuracy compared to the Spatial Transformer Network.

In the work of Montserrat et al.~\cite{montserrat2020deepfakes}, face forgery recognition in videos is proposed by using a weighting approach of the fake face probabilities in frames and a GRU layer for temporal learning of the frames' feature vectors. For each frame and face found in the video, the EfficientNet computes a weighted value and the logit value containing the probability of whether the face is real or fake in the feature map resulting from the CNN model. All weights and logit values are then used to compute the forgery likelihood $p_{w}$ for the entire video. The logit values, weights, and the final probability $p_{w}$ are concatenated to the feature vectors for further analysis by a GRU layer and final prediction as a real or deepfake video. This approach is called Automatic Face Weighting (AFW). The proposed method with AFW and GRU layer achieved the best accuracy of 91.88\% on the test set samples of the Deep Fake Detection Dataset (DFDC).

In a similar study, Hao et al.~\cite{hao2022deepfake} addressed detecting deepfake videos using a multimodality approach that relied upon visual and audio components of the video. For the visual classification, each video frame is fed to an EfficientNet-b5 CNN for feature extraction and further classification of the face as real or a possible forgery by manipulation. The labels assigned to each video frame are then used to determine the probability of possible manipulation of the entire video. Then, the frames' feature vectors and the associated probabilities are combined and fed to a GRU layer to capture spatiotemporal properties and predict the video as real or fake. The authors presented a simple approach for the audio classification in which audio signals' spectrograms are fed to a customized CNN architecture for the subsequent classification as real or fake.
Moreover, a multimodality approach based on audio and visual information from the video was also proposed to provide more discriminative features for the deepfake classification. Emotional features are also extracted from audio and visual components for further combination into a latent space of features for the final classification as a real or manipulated video. However, the authors did not present quantitative analysis or results achieved by the multimodality approach.

In the work of Jaiswal~\cite{jaiswal2021hybrid}, a hybrid model combining LSTM and GRU layers was proposed to exploit the benefits of each type of recurrent model in the deepfake classification of video frames. The author presented a deep learning architecture in which two layers of each recurrent model are stacked together, followed by a single dense layer for binary classification of a video as real or deepfake. A customized CNN architecture was stacked before the hybrid recurrent layers for temporal feature extraction from each video frame. The hybrid sequence of GRU layers followed by two LSTM layers attained the best accuracy against using only one type of recurrent model. The provided accuracy was 0.8165 for the GRU-LSTM layers in the Deep Fake Detection Challenge Dataset.

Tu et al.~\cite{tu2021deepfake} addressed the problem of deepfake detection by using a Convolutional GRU (ConvGRU) architecture for temporal learning of feature maps produced by a pre-trained Resnet50 CNN on a sequence of 10 video frames. ConvGRU was used because it is less complex and has less parameters than Convolutional LSTM (ConvLSTM). The proposed method attained 89.3\% AUC and 94.56\% accuracy on the celeb-DF(v2) dataset. The main drawback is the lack of important architecture information like the feature map's size resulting from both, ResNet50 and ConvGRU.

Ismail et al.~\cite{ismail2022integrated} proposed a hybrid approach for face forgey classification in videos that integrates image features extracted from a modified Xception Net architecture and spatial gradient directions computed from the Histogram of Gradient Oriented (HOG) method. Their strategy presented a customized CNN architecture that receives the image containing the gradient orientation calculated by the HOG method and produces a fixed-size output feature vector representation. Moreover, an improved Xception Net architecture is proposed to extract the feature's vector representation directly from the input video frames. The feature vectors produced by the two CNN models are then fused and fed to a sequence of GRU layers for further classification of the video's authenticity. To capture discontinuities produced by processing each frame individually, the authors utilized eight sequences of GRU layers to extract the temporal features of the video frames, which are then fed to a fully connected layer for the final video's classification as real or fake. The proposed method performed best compared to baseline CNNs, achieving 95.56\% accuracy and 95.53\% of Area Under the Receiver Operating Characteristic (AUROC) on the Celeb-DF and FaceForencics++ datasets.

Pu et al.~\cite{pu2022learning} proposed a temporal learning-based method and a novel loss function to handle deepfake detection in a class-imbalanced dataset. Using a video-level and a frame-level classification approach, the authors combined the feature maps extracted from 300 video frames with the temporal learning performed by GRU layers to classify real and fake faces in videos. ResNet50 has been used to compute the features from each video frame. Also, the authors proposed a loss function that combines the binary cross entropy and AUC to efficiently cope with imbalanced class distribution at the video-level and frame-level classification. Experiments were performed using the Celeb-DF and FaceForensics++ datasets. Also, the authors used samples from the DFDC dataset with different ratios of positive and negative instances to simulate an imbalanced data distribution. No data augmentation was used in this work. The proposed method attained the best performance at both the video-level and frame-level classification, even in skewed data distribution that promotes excessive samples for videos of real faces. The method achieved 96.5\% accuracy and 98.9\% AUC in the imbalanced Celeb-DF dataset. Also, the performance increased as the combined loss was included in the model.

Elpeltagy et al.~\cite{elpeltagy2023novel} addressed the ability of a multimodal-feature level approach for deepfake classification in videos. The proposed method is based on two modalities of features extracted from frames and the audio of the input videos. Each video component, i.e., the visual and the audio, is fed to a different CNN architecture to obtain two feature vector representations. The two feature vectors are then fused and given to a GRU network to learn the video's temporal properties. The real or fake video prediction is performed by a fully connected layer that receives the temporal features from the GRU model. From experiments performed on the FakeAVCeleb dataset, the method attained the highest accuracy (97.52\%) when compared to Xception and VGG16 employed for deepfake classification on spatial characteristics of the frames.

Sun et al.~\cite{sun23} designed a recent deepfake detection method to transform the task of detecting deep fake videos into a scheme of detecting multi-variable time series anomalies to expose artifacts generated by facial manipulation in both temporal and spatial dimensions. The authors propose employing virtual-anchor-based region displacement trajectory extraction to obtain the spatial-temporal representation of different facial areas. Furthermore, a fake trajectory detection network was constructed based on dual-stream spatial-temporal graph attention. A gated recurrent unit backbone converts the deep fakes detection task into a binary classification problem for a multi-variable time series. The samples from the Face-Forensics++ dataset were applied to carry out the method.

Table~\ref{t.gru} presents the summary of the methods described in this section, i.e., using GRU to deepfake detection.

\begin{center}
\scriptsize
\begin{longtable}{|C{.10\textwidth} | C{0.05\textwidth} | C{.13\textwidth} | C{.20\textwidth} | C{.13\textwidth} | L{.13\textwidth} |}
\caption{Summarized works considering GRU sorted by year and alphabetical order.} \label{t.gru} \\
	
	\hline
\multicolumn{6}{|c|}{\cellcolor{blue!10}\textbf{\small CNN+GRU}}  \\
	\hline \multicolumn{1}{|C{.10\textwidth} |}{\textbf{Ref.}} & \multicolumn{1}{C{0.05\textwidth}|}{\textbf{Year}} & \multicolumn{1}{C{.13\textwidth}|}{\textbf{Technique}} & \multicolumn{1}{C{.20\textwidth}|}{\textbf{Dataset}} & \multicolumn{1}{C{.13\textwidth}|}{\textbf{Input}} & \multicolumn{1}{C{.13\textwidth}|}{\textbf{Best result}} \\ \hline 
	\endfirsthead

\multicolumn{6}{c}%
{{\bfseries \tablename\ \thetable{} -- continued from previous page}} \\
\hline \multicolumn{1}{|p{.10\textwidth} |}{\textbf{Ref.}} & \multicolumn{1}{p{0.05\textwidth}|}{\textbf{Year}} & \multicolumn{1}{p{.13\textwidth}|}{\textbf{Technique}} & \multicolumn{1}{p{.20\textwidth}|}{\textbf{Dataset}} & \multicolumn{1}{p{.13\textwidth}|}{\textbf{Input}} & \multicolumn{1}{p{.13\textwidth}|}{\textbf{Best result}} \\ \hline 
\endhead

\hline \multicolumn{6}{|r|}{\cellcolor{gray!10}{Continued on next page}} \\ \hline
\endfoot
\hline \hline
\endlastfoot
\cite{sabir2019recurrent}   &    2019    &    DenseNet + Face Alignment &   FaceForensics++   &   Videos   &   Accuracy: 96.9\% \\ \hline
\cite{montserrat2020deepfakes}   &   2020   &   EfficientNet + AFW   &   DFDC  &  Videos   &   Accuracy: 91.88\% \\ \hline
\cite{jaiswal2021hybrid}   &   2021   &   CNN +  GRU-LSTM  &   DFDC  &  Videos  &  Accuracy: 81.65\% \\ \hline
\cite{tu2021deepfake}   &   2021 &  ConvGRU   &   Celeb-DF(v2)  &  Videos   &  Accuracy: 94.56\% \\ \hline
\cite{hao2022deepfake}   &   2022   &    EfficientNet-b5   &   DFDC   &   Videos   &   AUC: 0.97 \\ \hline
\cite{ismail2022integrated}  &  2022  &  XceptionNet + HOG   &   Celeb-DF and FaceForensics++  &  Videos  &  Accuracy: 95.56\% \\ \hline
\cite{pu2022learning}  &   2022   &   ResNet-50   &   Celeb-DF  &  Videos   &   Accuracy: 96.5\% \\ \hline
\cite{elpeltagy2023novel}   &   2023   &   XceptionNet + InceptionResNet   &   FakeAVCeleb  &   Videos and Audio  &   Accuracy: 97.52\% \\ \hline
\cite{sun23}   &   2023  &  Trajectory of the Facial Region Displacement  &  FaceForensics++ & Video & Accuracy: 99.5\%\textsuperscript{$\ast$} \\ \hline

\multicolumn{6}{l}{\textsuperscript{$\ast$}Maximum score obtained from the deepfake manipulation method of the FaceForensics++.}\\
\end{longtable}
\end{center}

\subsection{Transformers}

Khan et al.~\cite{hang21} propose a video transformer with a face UV Texture Map for deepfake detection. The results on five public datasets show that the method achieves better than state-of-the-art methods. That proposed segment embedding allows the network to extract more informative features, improving detection accuracy. The exhaustive experiments show that the model can reach suitable performance on an unexplored dataset while maintaining the performance on the previous dataset.

Coccomini et al.~\cite{tombari22} investigate various solutions based on combinations of convolutional networks, mainly the EfficientNet-B0, with varying types of Vision Transformers and compare the results with the state-of-the-art. The proposed solution is designed to merge two visual transformer architectures which combine multi-scaled feature maps obtained by two pre-trained EfficientNet-B0 CNNs. By combining feature representations of the transformer mechanism, the method can learn the deepfake aspects of the multi-scale feature representation of the face. Still is investigating some gains that can be made during generalization to achieve better and more stable results in video deepfake detection. The work employs a patch extractor based on EfficientNet. It is particularly effective even just using the smallest network in this category. It led to better outcomes than an extractor with a generic convolutional network trained from scratch, thus achieving an AUC of 0.951 using the cross-visual transformer. Moreover, compared to state-of-the-art solutions, the method achieved the highest mean accuracy in the four face manipulation strategies of the FaceForensics++ dataset.

Heo et al.~\cite{heo23} proposed a DeepFake detection using a Vision Transformer Model, which has indicated good performance in recent image classifications and combined CNN and patch-embedding features during the input stage. The Robust Vision Transformer Model has shown efficiency compared with EfficientNet as the state-of-the-art model, which consists of a 2D CNN network. The state-of-the-art obtained an AUC of 0.972, whereas the proposed work obtained 0.978 under identical conditions without an ensemble approach. The proposed method produced an F1 score of 0.919, whereas the state-of-the-art model achieved 0.906 under the same threshold condition of 0.55. Furthermore, the authors observed an AUC gain of up to 0.17 compared with a recent scheme. The proposed model reached an AUC of 0.982 with the ensemble method, whereas the state-of-the-art model achieved 0.981.

The work developed by Xue et al.~\cite{xue22} proposes a transformer-based deepfake detection method for facial organs, which can effectively differentiate deepfake media. The authors highlight that deepfake detection on subtle-expression manipulation, facial-detail modification, and smeared images has become a wide research hotspot. Also, complete that existing deepfake detection methods on the entire face are coarse-grained, where the details are missing due to the insignificant manipulated size of the image. To address the concerns, the authors created a transformer model for a deepfake detection method by organ. The investigation reduces the detection weight of defaced or unclear organs to prioritize the detection of clear and undamaged organs. The study also implements a Facial Organ Forgery Detection Test Dataset (FOFDTD), which includes the images of the masked face, sunglasses face, and undecorated faces collected from the network. Experimental results verified the effectiveness of the proposed approach, which attained an AUC of 99.93\%, 94.32\%, 75.93\%, and 82.43\% in the FaceForensics++, DFD, DFDC-P, and Celeb-DF datasets, respectively.

In their paper, Zhang et al.~\cite{zhang2022robust} propose the TransDFD, a transformer-based network for deepfake detection that learns discriminative and general manipulation patterns end-to-end. Their model introduces the spatial attention scaling module, which emphasizes salient features while suppressing less important representations. It considers fine-grained local and global features based on intra-patch locally-enhanced relations. Additionally, it also finds inter-patch locally-enhanced global relationships in face images. Experiments conducted over several public benchmark datasets show that TransDFD can outperform state-of-the-art approaches in robustness and computational efficiency.

The work of Khan et al.~\cite{khan2022hybrid} proposes a hybrid transformer network using a feature fusion strategy for deepfake video detection. The model employs XceptionNet and EfficientNet-B4 as feature extractors along with a transformer architecture in an end-to-end manner on FaceForensics++ and DFDC benchmarks. The authors also proposed two augmentation techniques: face cut-out and random cut-out augmentations. The model achieved comparable results to more advanced state-of-the-art approaches, while the augmentation techniques improved the detection performance of the model and reduced overfitting.

Khormali et al.~\cite{khormali2022dfdt} proposed an end-to-end Transformers-based deepfake detection framework called DFDT, whose layers implement a re-attention mechanism instead of a traditional multi-head self-attention layer. The model learns hidden traces of perturbations from local image features and the global relationship of pixels at different forgery scales using four main components: patch extraction and embedding, multi-stream transformer block, attention-based patch selection, and a multi-scale classifier. The performance of the approach is accessed through a set of experiments on several deepfake forensics benchmarks, which results reached detection rate values of $99.41\%$, $99.31\%$, and $81.35\%$ on FaceForensics++, Celeb-DF (V2), and WildDeepfake, respectively.

Coccomini et al.~\cite{coccomini2022cross} considered the possibility of untying the deepfake detection to the methods used to generate the training samples. The authors compared Vision Transformer with an EfficientNetV2 on a cross-forgery context based on the ForgeryNet dataset~\cite{He2021ForgeryNet}, concluding that EfficientNetV2 has a greater tendency to specialize, often obtaining better results on training methods. At the same time, Vision Transformers exhibit a superior generalization ability, making them competent even on images generated with new methodologies.

The work of Wang et al.~\cite{wang2022m2tr} proposed the Multi-modal Multi-scale TRansformer (M2TR), which aims to capture subtle manipulation artifacts at different scales using transformers. The model operates on patches of different sizes to detect local inconsistencies in images at different spatial levels, also learning to detect forgery artifacts in the frequency domain to complement RGB information through a cross-modality fusion block. Results show that the technique can outperform state-of-the-art deepfake detection methods by clear margins when applied to a novel large-scale deepfake dataset named Swapping and Reenactment DeepFake (SR-DF).

A more recent work~\cite{wang2023deep} proposes a deep convolutional Transformer model that incorporates decisive image features locally and globally. The model applies convolutional pooling and re-attention to enrich the extracted features and image keyframes to improve the deepfake detection performance and visualize the feature quantity gap between the key and normal image frames caused by video compression. The experiments conducted over several deepfake benchmark datasets show that the solution outperforms several state-of-the-art baselines considering both within- and cross-datasets.

Raza, Malik and Haq~\cite{raza2023holisticdfd} propose a vision transformer architecture combining spatial, temporal and spatiotemporal features extracted from videos for deepfake classification tasks. Spatial feature extraction is achieved by two-dimensional convolutional layers applied to a single frame of the video. In contrast, temporal features are extracted using three-dimensional convolutions to a sequence of images comprising the difference between two consecutive video frames. Finally, 3D convolutions are applied directly to video frames to capture the spatiotemporal aspects of the face. The proposed strategy combines the transformer representations obtained from the spatial, temporal, and spatiotemporal feature maps into a single feature vector representation which is then fed to a fully connected layer. This strategy can capture evidence of possible manipulations at different feature levels, i.e., spatial and temporal domains. Results show AUC scores of 0.926, 0.9624, and 0.9415 on the DFDC, Celeb-DF, and FaceForensics++ datasets, respectively. Moreover, the proposed method achieved the best accuracy in videos produced by the Neural Texture subset of the FaceForensics++ dataset.

Feinland et al.~\cite{feinland2022poker} proposed merging two visual transformer architectures to combine multi-scaled feature maps obtained by two pre-trained EfficientNet-B0 CNNs. By combining feature representations in the attention mechanism, the method can learn important aspects at the multi-scale feature level of the face image. Moreover, the authors propose an inference approach ruled by the vote of predictions produced from each face detected per person in the video. The entire video is considered a forgery if one person's face is classified as fake. The method attained an AUC of 0.951 using the voting classification and the cross-visual transformer with EfficentNet-B0 as the backbone for feature extraction. Compared to state-of-the-art methods, the approach achieved the highest mean accuracy in the four face manipulation strategies of the FaceForensics++ dataset.

The work of Lin et al.~\cite{lin2023deepfake} proposes a dual-subnet network that uses a transformer architecture to learn and extract multi-scale information and high-level features of the faces to cope with deepfake in videos. Using multi-scale information makes it possible to learn intrinsic aspects revealing possible manipulations at different regions of the target face. At the same time, high-dimensional features are extracted via an EfficientNet-B4 convolutional module with depthwise convolutions. The multi-scale and the high-dimensional features are combined and fed to a vision transformer module to learn more contextual relations among the image features, followed by the final classification of the video as real or fake. By exploiting features at different scales, the method achieved the best scores in all datasets and ablation scenarios, with the best accuracy of 99.80\% on the Celeb-DF dataset. In comparison, the worst performance was attained on the WildDeepfake dataset (82.63\%).

Zhang et al.~\cite{zhang2022deepfake} employed a vision transformer architecture for the temporal analysis of faces' random regions to cope with spatiotemporal inconsistencies indicating possible video manipulation. The method is called spatiotemporal dropout, which discards some facial frames and random patches of each frame based on a uniform distribution ruled by dropout rates. A bag of patches is then formed from the selected facial regions and fed to the vision transformer architecture to capture inconsistencies across the frames. A fully connected layer then uses the transformer representation to predict the video as real or fake. Since the counterfeit artifacts are mostly spread across some regions of the face, the model can capture more specific features which locally describe spatial inconsistencies. Results showed the best AUC scores compared to 25 state-of-the-art methods in all the deepfake datasets, achieving average scores of 99.8\%, 99.1\%, and 97.2\% in the FaceForensics++, DFDC, and Celeb-DF datasets, respectively. Moreover, the model was able to cope with the four facial manipulations of the FaceForensics++ dataset, thus achieving eminent performance with scores higher than 90\% in all subsets of deepfake generation.

Khalid, Akbar, and Gul~\cite{Khalid:2023} created the Swin Y-Net Transformers architecture in which the encoder, composed of a swin transformer, divides the entire image into patches to extract details. In contrast, the decoder, composed of U-Net, creates a segmentation mask for further classification. Experiments conducted over Celeb-DF and FF++ datasets demonstrated the generalization capability of the proposed model and great capacity to identify videos created by DeepFakes, FaceSwap, Face2Face, FaceShifter, and NeuralTextures algorithms.

Table~\ref{t.transformers} presents the summary of the methods described in this section, i.e., using Transformers to deepfake detection.

\begin{center}
\scriptsize
\begin{longtable}{|C{.10\textwidth} | C{0.05\textwidth} | C{.13\textwidth} | C{.20\textwidth} | C{.13\textwidth} | L{.13\textwidth} |}
\caption{Summarized works considering Transformers sorted by year and alphabetical order.} \label{t.transformers} \\
	
	\hline
\multicolumn{6}{|c|}{\cellcolor{blue!10}\textbf{\small Transformers}}  \\
	\hline \multicolumn{1}{|C{.10\textwidth} |}{\textbf{Ref.}} & \multicolumn{1}{C{0.05\textwidth}|}{\textbf{Year}} & \multicolumn{1}{C{.13\textwidth}|}{\textbf{Technique}} & \multicolumn{1}{C{.20\textwidth}|}{\textbf{Dataset}} & \multicolumn{1}{C{.13\textwidth}|}{\textbf{Input}} & \multicolumn{1}{C{.13\textwidth}|}{\textbf{Best result}} \\ \hline 
	\endfirsthead

\multicolumn{6}{c}%
{{\bfseries \tablename\ \thetable{} -- continued from previous page}} \\
\hline \multicolumn{1}{|p{.10\textwidth} |}{\textbf{Ref.}} & \multicolumn{1}{p{0.05\textwidth}|}{\textbf{Year}} & \multicolumn{1}{p{.13\textwidth}|}{\textbf{Technique}} & \multicolumn{1}{p{.20\textwidth}|}{\textbf{Dataset}} & \multicolumn{1}{p{.13\textwidth}|}{\textbf{Input}} & \multicolumn{1}{p{.13\textwidth}|}{\textbf{Best result}} \\ \hline 
\endhead

\hline \multicolumn{6}{|r|}{\cellcolor{gray!10}{Continued on next page}} \\ \hline
\endfoot
\hline \hline
\endlastfoot
\cite{khan2022hybrid}   &    2022    &    UV Texture Map   &   FaceForensics++ and DFDC  &  Video   &  Accuracy: 99.79\%\textsuperscript{\ddag} \\ \hline
\cite{tombari22}   &    2022   &   EfficientNet-B0    &   FaceForensics++ and DFDC   &   Video   &   AUC: 0.951\textsuperscript{\ddag} \\ \hline
\cite{feinland2022poker}   &    2022    &   EfficientNet-B0   &  FaceForensics++ and DFDC  &  Videos   &   AUC: 0.951 \\ \hline
\cite{xue22}   &   2022    &  CNN on multiple face organs  &    FaceForensics++, DFD, DFDC-P and Celeb-DF   &   Video   &   AUC: 99.93\%\textsuperscript{\ddag} \\ \hline
\cite{zhang2022robust}   &   2022    &   VGG  &   FaceForensics++, DFDC and DFD    &   Video   &   AUC: 98.40\% \\ \hline
\cite{khan2022hybrid}   &    2022    &     Face cut-out and Random cut-out   &  FaceForensics++ and DFDC &   Video    &    Accuracy: 98.24\%\textsuperscript{\ddag} \\ \hline
\cite{khormali2022dfdt}  &   2022   &  Patch extraction and embedding   &   FaceForensics++, Celeb-DF and WildDeepfake  &  Videos   &   Accuracy: 99.41\%\textsuperscript{\ddag} \\ \hline
\cite{coccomini2022cross}   &   2022    &  Vision Transformer  &   ForgeryNet~\cite{He2021ForgeryNet}  &  Images  &   Variance: 0.004   \\ \hline
\cite{wang2022m2tr}   &    2022    &  CNN + Frequency Filter &  FaceForensics++, Celeb-DF and SR-DF\textsuperscript{\ddag}  &    Videos   &    AUC: 91.20\%\textsuperscript{$\S$} \\ \hline
\cite{zhang2022deepfake}    &    2022    &   Dropout rate to discard image patches    &    Celeb-DF, DFDC and FaceForensics++ &   Videos   &   AUC:  99.8\%\textsuperscript{\ddag} \\ \hline
\cite{raza2023holisticdfd}   &    2023   &    2D and 3D CNNs   &    Celeb-DF, DFDC and FaceForensics++   &   Videos  &   AUC: 0.9624\textsuperscript{\ddag} \\ \hline
\cite{heo23}   &   2023   &   EfficientNet-B7   &   DFDC and Celeb-DF (v2)   &    Video    &     AUC: 0.982 \\ \hline
\cite{lin2023deepfake}   &    2023    &    EfficientNet-B4   &  Celeb-DF, DFDC, Face-Forensics++, and WildDeepfake & Videos    &     Accuracy: 99.80\%\textsuperscript{\ddag} \\ \hline
\cite{Khalid:2023}   &    2023   &   Encoder + Decoder - Transformer   &    Celeb-DF and FaceForensics++   &    Videos   &    AUC: 0.99\textsuperscript{\ddag} \\ \hline
\cite{wang2023deep}   &   2023   &    Local and global feature maps   &    FaceForensics++, Celeb-DF, DF-1.0 and DFDC   &   Videos   &    AUC: 97.66\% \\ \hline

\multicolumn{6}{l}{\textsuperscript{\ddag}Maximum score when the specified datasets are tested individually.}\\
\multicolumn{6}{l}{\textsuperscript{$\S$}Score obtained from the novel SR-DF dataset used in the training and testing of the model.}
\end{longtable}
\end{center}

\section{Datasets}
\label{s.datasets}

This section presents the most recent and popular datasets generated with deep learning techniques for deepfake detection.

\subsection{HOHA-based dataset}
\label{ss.hoha}

G{\"u}era and Delp~\cite{guera2018deepfake} provided a dataset composed of $300$ videos randomly selected from the HOHA dataset~\cite{laptev2008learning}, which comprises a realistic set of sequence samples from famous movies with an emphasis on human actions, as well as $300$ other deepfake videos collected from multiple video-hosting websites, leading to a total of $600$ videos, usually presented in $360\times240$ format, with $24$ frames per second.

\subsection{Faceswap-GAN}
\label{ss.faceswapGAN}

\begin{figure*}[!htb]
    \centering
    \includegraphics[width=.96\textwidth]{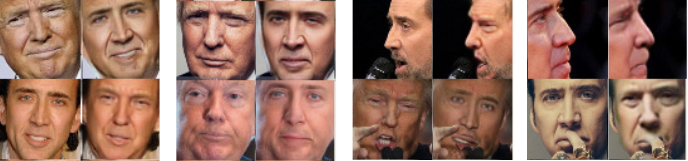}
    \caption{Samples from Faceswap-GAN dataset. For each block, the left column denotes original images and the right column stands for synthetic instances. Adapted from~\cite{faceswapGAN}.}
    \label{f.faceswap}
\end{figure*}

Korshunov and Marcel~\cite{korshunov2019vulnerability} proposed the first publicly dataset composed of deepfake videos created with GANs, i.e., the Faceswap-GAN database\footnotemark. The dataset comprises low and high-quality videos with $64\times64$ and $128\times 128$ pixels resolution, respectively. Each resolution comprises $320$ samples with approximately $200$ frames each. Finally, it is generated from $16$ pairs of people manually selected from the VidTIMIT dataset. Figure~\ref{f.faceswap} depicts some examples.

\footnotetext{https://github.com/shaoanlu/faceswap-GAN}

\subsection{UADFV}
\label{ss.uadfv}

\begin{figure*}[!htb]
    \centering
    \includegraphics[width=.96\textwidth]{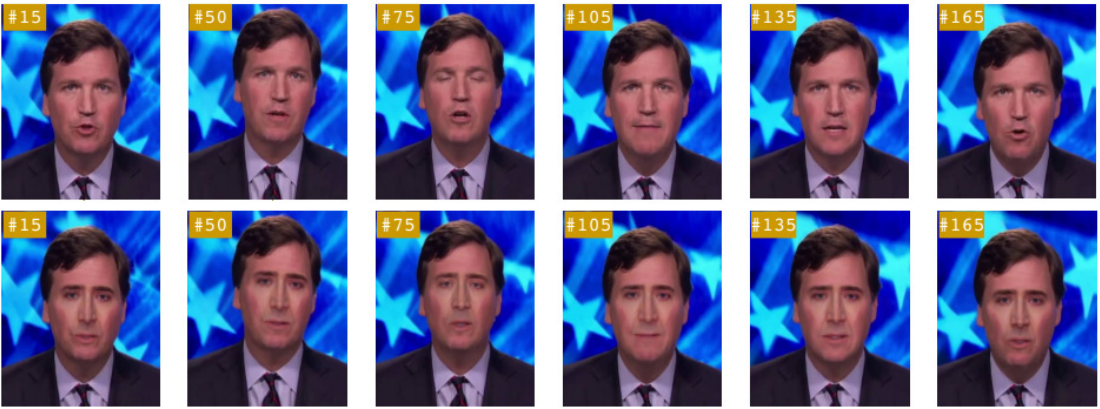}
    \caption{Sample frames from the UADFV dataset. The top row depicts original faces, while the bottom row stands for synthetic images. Adapted from~\cite{li2018ictu}.}
    \label{f.uadfv}
\end{figure*}

The UADFV~\cite{li2018ictu} is a synthetic dataset provided by the University of Albany with the primary objective of helping to detect fake face videos through physiological signals, i.e., eye blinking, a feature claimed by the authors as not well presented in synthesized videos. The dataset is composed of $49$ fake videos generated through the FakeApp mobile application~\footnote{https://fakeapp.softonic.com}, where the individual's original faces are swapped with Nicolas Cage's face. Each sequence comprises a $294\times500$ pixels resolution and $11.14$ seconds on average. Figure~\ref{f.uadfv} provides some samples of the original and their respective synthetic version.

\subsection{Deepfake-TIMIT}
\label{ss.deepfaketimit}

\begin{figure}[!htb]
    \centering
     \begin{tabular}{cccc}
      \includegraphics[width=.21\textwidth]{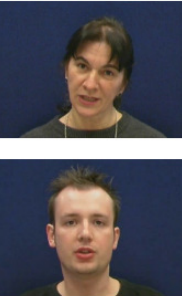} &
      \includegraphics[width=.21\textwidth]{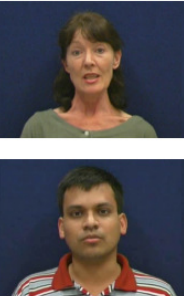} &
      \includegraphics[width=.21\textwidth]{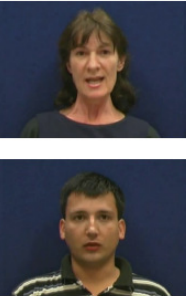} &
      \includegraphics[width=.21\textwidth]{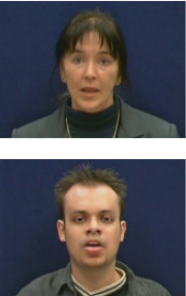} \\
      (a) & (b) & (c) & (d)
     \end{tabular}
    \caption{Sample frames from the Deepfake-TIMIT dataset: (a) original image $A$, (b) original image $B$, (c) swap $A\rightarrow B$, and (d) swap $B\rightarrow A$. Adapted from~\cite{korshunov2018deepfakes}.}
    \label{f.timit}
\end{figure}

The Deepfake-TIMIT~\cite{korshunov2018deepfakes} comprises $640$ fake videos obtained from $10$ image sequences of $32$ people extracted from VidTIMIT dataset~\footnote{https://conradsanderson.id.au/vidtimit/}, generated using a GAN-based face-swapping algorithm. The authors manually selected $16$ pairs of individuals that shared some visual similarities and swapped their faces, as illustrated in Figure~\ref{f.timit}. The videos are divided into two main categories, i.e., low quality, which comprises $320$ videos with approximately $200$ frames of $64\times64$ pixels each, and high quality, composed of $320$ image sequences with around $400$ frames of size $128\times128$ pixels. 

\subsection{FaceForensics}
\label{ss.faceforensics}

\begin{figure}[!htb]
    \centering
     \begin{tabular}{cccc}
      \includegraphics[width=.21\textwidth]{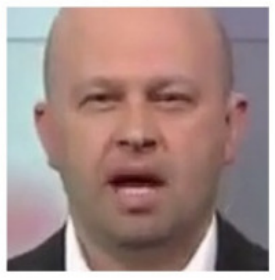} &
      \includegraphics[width=.21\textwidth]{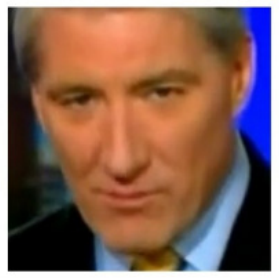} &
      \includegraphics[width=.20\textwidth]{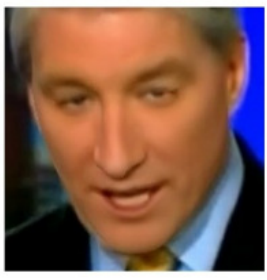} &
      \includegraphics[width=.21\textwidth]{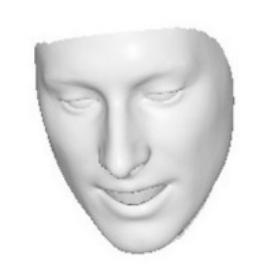} \\
      (a) & (b) & (c) & (d)
     \end{tabular}
    \caption{Faceforensics reenactment example: (a) original (source), (b) original (target), (c) manipulated, and (d) mask. Adapted from~\cite{rossler2018faceforensics}.}
    \label{f.faceforensics}
\end{figure}

FaceForensics dataset~\cite{rossler2018faceforensics} comprises about a half-million manipulated images from $1,004$ videos designed for benchmarking forensic purposes regarding classification and segmentation tasks at various quality levels, provided with ground-truth masks. The dataset is divided into two subsets created using Face2Face~\cite{thies2016face2face} reenactment approach such that the first, namely Source-to-Target Reenactment Dataset, performs the reenactment between two randomly chosen videos, as illustrated in Figure~\ref{f.faceforensics}, and the second subset is the Self-Reenactment Dataset, which uses the same video as the source and the target video. The authors considered videos with $854\times480$ resolution or more from the youtube and youtube8m datasets and extracted sequences containing at least $300$ consecutive frames of the face. Finally, manual screening is performed to assure the videos' quality. The whole dataset comprises $1,408$ videos for training, $300$ for validation, and $300$ for testing purposes, resulting in $732,391$, $151,835$, and $156,307$ images, respectively.  

\subsection{Faceforensics++}
\label{ss.faceforensicsplusplus}

\begin{figure}[!htb]
    \centering
     \begin{tabular}{cccc}
      \includegraphics[width=.47\textwidth]{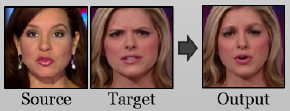} &
      \includegraphics[width=.47\textwidth]{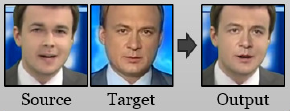} \\
      (a) & (b) 
     \end{tabular}
    \caption{Examples of Faceforensics++ approaches for face reenactment (a) and replacement (b). Adapted from~\cite{rossler2019faceforensics}}
    \label{f.forensics++}
\end{figure}

FaceForensics++~\cite{rossler2019faceforensics} is an extension of the FaceForensics dataset and denotes a public dataset proposed as a benchmark for realistic fake face image detection. The set comprises $1,000$ thoroughly selected videos, most of them from YouTube, such that approximately $60\%$ of the individuals are male and the remaining $40\%$ are female. Concerning the resolution, approximately $55\%$ are provided with $854\times480$, i.e., Video Graphics Array (VGA) resolution, $32,5\%$ in $1,280\times720$, i.e., high definition (HD), and $12,5\%$ in $1,920\times1,080$ (full-HD) resolutions. Further, the authors performed a manual screening to ensure high-quality and avoid face occlusion, and exposed the videos to four face manipulation approaches, i.e., NeuralTextures~\cite{thies2019deferred}, Face2Face~\cite{thies2016face2face}, FaceSwap~\cite{thies2016face2face}, and Deepfakes\footnote{\url{https://github.com/deepfakes/faceswap}}. As an output, the model provides a manipulated video and a ground-truth mask indicating modified pixels for each input video to provide a more robust training data set. Figure~\ref{f.forensics++} illustrates examples of face reenactment and replacement present in Faceforensics++ dataset. 

\subsection{Deepfake Detection Challenge}
\label{ss.facebook}

\begin{figure}[!htb]
    \centering
      \includegraphics[width=.90\textwidth]{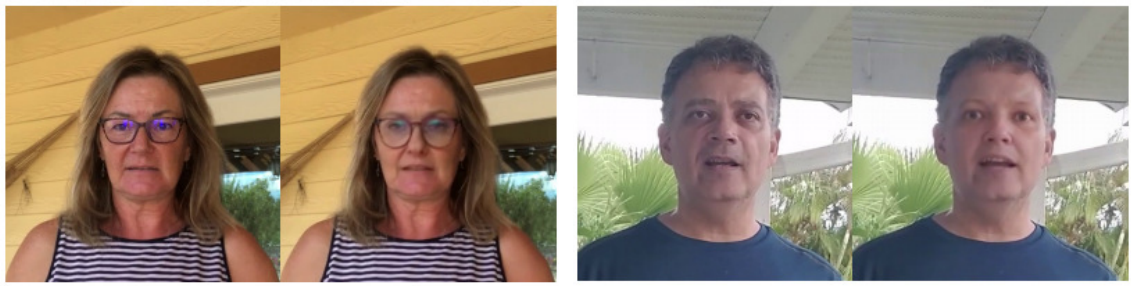} 
    \caption{Examples from DFDC dataset. Adapted from~\cite{dolhansky2019deepfake}.}
    \label{f.dfdc}
\end{figure}

Facebook's Deepfake Detection Challenge\footnote{\url{https://ai.facebook.com/datasets/dfdc/}} (DFDC)~\cite{dolhansky2019deepfake} dataset consists of $5,000$ videos from actors with face likenesses manipulated. The dataset comprises $66$ actors selected respecting the following proportions: $26\%$ male and $74\%$ female, $3\%$ south-Asian, $9\%$ west-Asian, $20\%$ African-American, and $68\%$ Caucasians. The manipulation was conducted considering two face swap approaches: method A, which produces higher swap quality images with faces closer to the camera, considering the source and swapped faces in the same proportions, and method B, which has lower quality swaps. In the end, we have a dataset composed of $4,464$ sample clips for training purposes and $780$ for testing, each with $15$ seconds length and different resolutions. Figure~\ref{f.dfdc} provides some examples of the dataset.

\subsection{Celeb-DF}
\label{ss.celebDF}

\begin{figure*}[!htb]
    \centering
    \includegraphics[width=.96\textwidth]{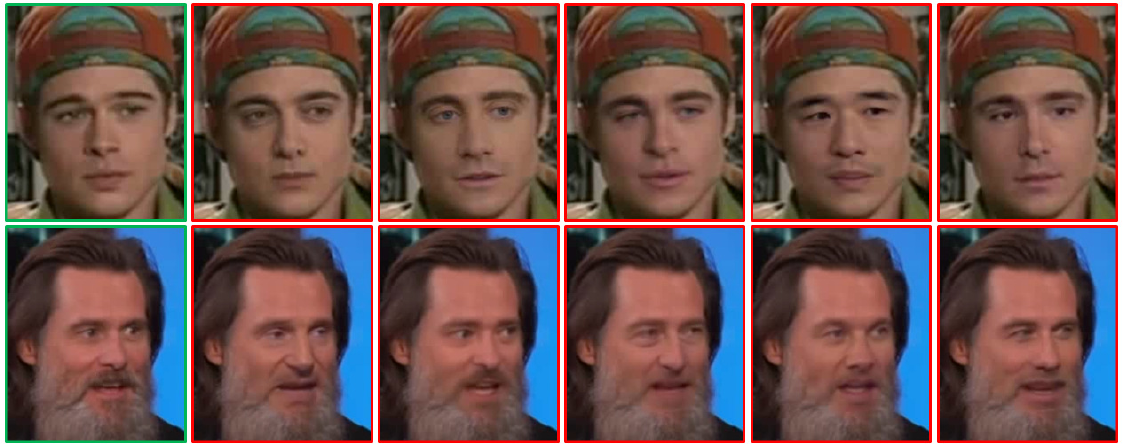}
    \caption{Celeb-DF dataset samples. Green-framed instances denote real images, while the red-framed ones stand for the corresponding fake samples generated through random donor individuals.}
    \label{f.deleb-df}
\end{figure*}

Celeb-DF~\cite{li2020celeb} is a challenging large-scale deepfake video dataset generated using an improved synthesis process over celebrities' videos available on YouTube. The dataset comprises $5,639$ high-quality videos with more than two million frames of size $256\times256$ pixels each from $59$ celebrities, comprising diverse ethnic groups ($88.1\%$ are Caucasians, $5.1\%$ are Asians, and $6.8\%$ are African Americans), ages ($6.4\%$ under $30$ years, $28.0\%$ between $30$ and $40$, $26.6\%$ are $40$s, $30.5\%$ between $50$ and $60$, and $8.5\%$ are of age $60$ or above), and genders ($56.8\%$ male and $43.2\%$ female). Each video has approximately $13$ seconds with a standard frame rate of $30$ frames-per-second and depicts diverse aspects such as lighting conditions, orientations, backgrounds, and subjects' face sizes (in pixels). Figure~\ref{f.deleb-df} provides some dataset examples. 

\subsection{DeeperForensics-1.0}
\label{ss.deeperforensics}

\begin{figure}[!htb]
    \centering
     \begin{tabular}{cccc}
      \includegraphics[width=.47\textwidth]{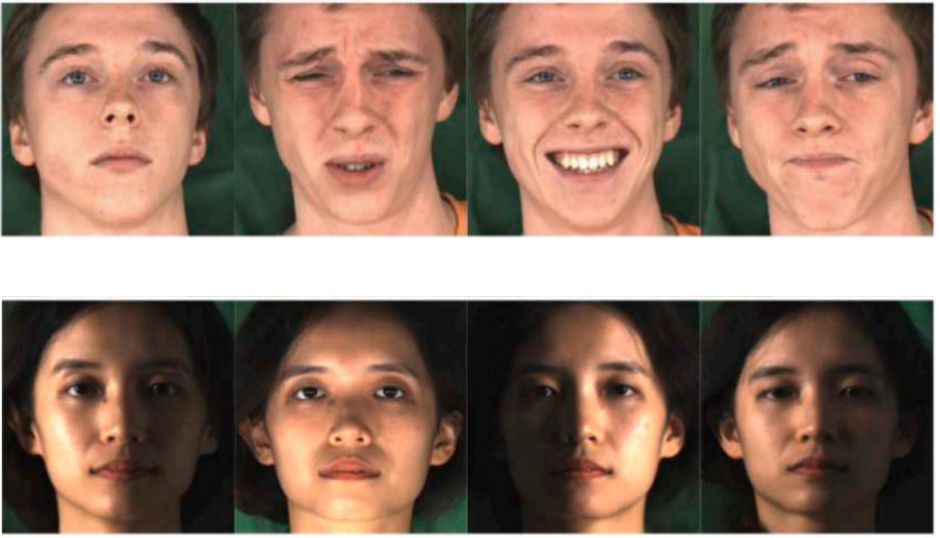} &
      \includegraphics[width=.47\textwidth]{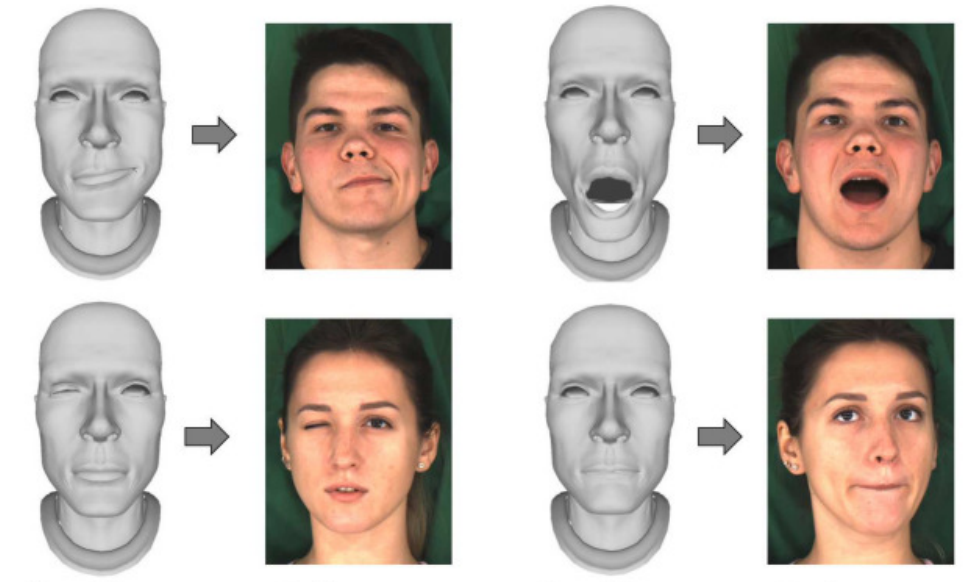} \\
      (a) & (b) 
     \end{tabular}
    \caption{Samples from DeeperForensics-1.0 considering: (a) expressions and different lighting conditions for the top and bottom rows, respectively, and (b) 3DMM blendshapes simulations. Adapted from~\cite{jiang2020deeperforensics}.}
    \label{f.deeperForensics}
\end{figure}

DeeperForensics-1.0~\cite{jiang2020deeperforensics} is large-scale, high-quality, and rich-diversity dataset designed for forgery detection. It comprises $60,000$ videos with $1,920\times 1,080$ resolution, comprising $17.6$ million frames of automatically generated swapped faces. The source videos were collected from $100$ actors from $26$ countries, distributed among males and females ranging from $20$ to $45$ years-old and diverse skin tones. Additionally, they were requested to perform eight natural expressions, i.e., fear, disgust, anger, happiness, contempt, surprise, sadness, and neutral, in distinct angles ranging from $-90^o$ to $+90^o$, and simulated $53$ other expressions from 3DMM blendshapes~\cite{cao2013facewarehouse}. The dataset considers variations in the video footage to match real-world cases, such as transmission errors, compression, and blurry. Besides, it also provides special attention to expressions, poses, and lighting conditions on source images since they perform a critical role in the dataset's quality. Figure~\ref{f.deeperForensics}(a) illustrates some examples of expressions (top row) and different lighting conditions (bottom row), while Figure~\ref{f.deeperForensics}(b) depicts some examples of 3DMM blendshapes simulations.

\subsection{Real and Fake Face Detection}
\label{ss.realAndFakeFace}

\begin{figure}[!htb]
    \centering
     \begin{tabular}{cccc}
      \includegraphics[width=.21\textwidth]{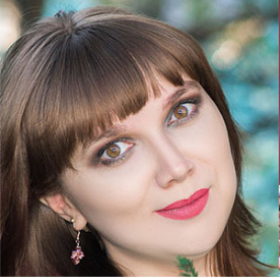} &
      \includegraphics[width=.21\textwidth]{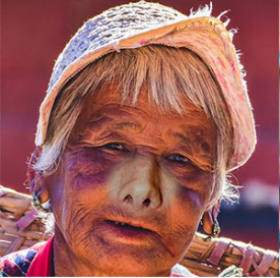} &
      \includegraphics[width=.21\textwidth]{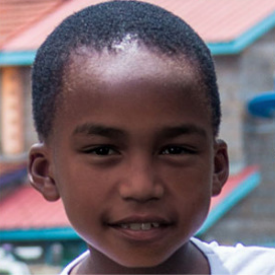} &
      \includegraphics[width=.21\textwidth]{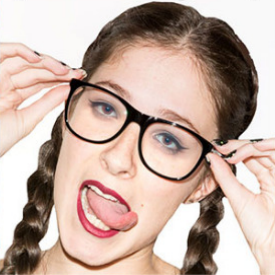} \\
      (a) & (b) & (c) & (d)
     \end{tabular}
    \caption{Examples: (a) real and fake examples in (b) easy (nose), (c) mid (face), and (d) hard (both eyes). Adapted from~\cite{realAndFake:2019}.}
    \label{f.realAndFakeFace}
\end{figure}

Real and Fake Face Detection\footnote{\url{https://www.kaggle.com/ciplab/real-and-fake-face-detection}.}~\cite{realAndFake:2019} is a dataset created by the Computational Intelligence and Photography Lab from Yonsei University, which comprises high-quality photoshopped face images. The main idea of using expert-generated images instead of generative models is to provide an alternative dataset for forged faces with a completely different set of features. The authors claim that a classifier trained using deepfakes can learn intrinsic patterns between real and GAN-generated images. On the other hand, such patterns are not present in experts' designs, creating counterfeits in a completely different process. The dataset figures three categories, i.e., easy, mid, and hard. Moreover, it comprises $1,081$ real and $960$ fake images of size $600\times600$ pixels. Figure~\ref{f.realAndFakeFace} illustrates some dataset examples.

\subsection{WildDeepfake}
\label{ss.wildDeepfake}

\begin{figure}[!ht]
    \centering
     \begin{tabular}{cccc}
      \includegraphics[width=.21\textwidth]{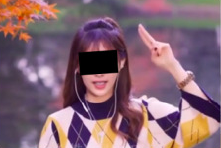} &
      \includegraphics[width=.21\textwidth]{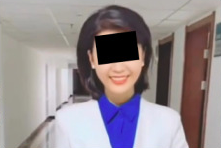} &
      \includegraphics[width=.21\textwidth]{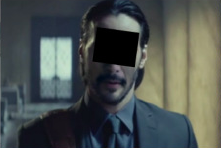} &
      \includegraphics[width=.21\textwidth]{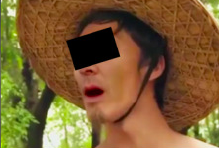} \\
      (a) & (b) & (c) & (d)
     \end{tabular}
    \caption{WildDeepfake dataset samples. The images comprise scene diversity to provide more realistic and real-world-like challenging scenarios. The authors block the eye regions due to privacy concerns. Adapted from~\cite{zi2020wilddeepfake}.}
    \label{f.wildDeepFake}
\end{figure}

WildDeepfake\footnote{\url{https://github.com/deepfakeinthewild/deepfake-in-the-wild}.} dataset~\cite{zi2020wilddeepfake} was proposed to better support real-world deepfake detection. The authors claim that deepfake datasets are usually filmed with a limited number of actors and scenes, and the videos are crafted using a few deepfake software, which impacts reduced effectiveness when detecting fake videos in the wild. In this context, WildDeepfake comprises $7,314$ face sequences of real and deepfake videos extracted from various sources on the Internet to provide a wide diversity of individuals, poses, and backgrounds. The data is divided into $6,508$ sequences for training and $806$ for testing purposes. Figure~\ref{f.wildDeepFake} depicts some dataset examples.

\subsection{Fake Face in the Wild }
\label{ss.fakeFaceWild}

\begin{figure}[!ht]
    \centering
     \begin{tabular}{cccc}
      \includegraphics[width=.21\textwidth]{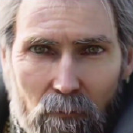} &
      \includegraphics[width=.21\textwidth]{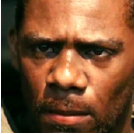} &
      \includegraphics[width=.21\textwidth]{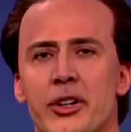} &
      \includegraphics[width=.21\textwidth]{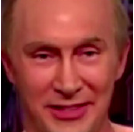} \\
      (a) & (b) & (c) & (d)
     \end{tabular}
    \centering
     \begin{tabular}{cccc}
      \includegraphics[width=.21\textwidth]{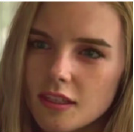} &
      \includegraphics[width=.21\textwidth]{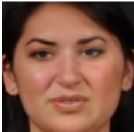} &
      \includegraphics[width=.21\textwidth]{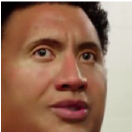} &
      \includegraphics[width=.21\textwidth]{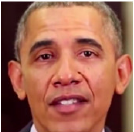} \\
      (e) & (f) & (g) & (h)
     \end{tabular}
    \caption{FFW dataset samples. Images (a) and (b) denote CGI full scenes, while (c) and (d) stand for deepfakes. Image (e) stands for head CGI, (f) represents face replacement, (g) denotes face CGI, and (h) represents part of face splicing. Adapted from~\cite{khodabakhsh2018}.}
    \label{f.ffw}
\end{figure}

The Fake Face in the Wild (FFW) dataset~\cite{khodabakhsh2018} tries to simulate the performance of fake face detection methods in the wild. The dataset figures $150$ videos extracted from YouTube. The selected videos denote fake content digitally created using GANs and CGI and manual and automatic image tampering techniques and their combinations. Moreover, each video length ranges from $2$ to $74$ seconds, with $854\times480$ resolution and $30$ frames per second, ending up in $53,000$ images. Figure~\ref{f.ffw} provides some examples of the dataset.

\subsection{Dataset Summary}
\label{ss.datasetsSummary}

Table~\ref{t.deepfake} introduces a summary of the datasets presented in this section.

\begin{table}[!htb]
\caption{Summary of the datasets used for deepfake detection.} \label{t.deepfake}
\centering
\resizebox{\textwidth}{!}{\begin{tabular}{|C{1cm}|L{3cm}|c|c|L{2.6cm}|C{2cm}|}
\hline
\multicolumn{6}{|c|}{\cellcolor{gray!20}{\large\textbf{Datasets}}} \\
\hline \textbf{Ref.} & \textbf{Dataset name} & \textbf{Year} & \textbf{Modalities} & \textbf{\# examples} & \textbf{Last works}\\ \hline


\cite{guera2018deepfake} & HOHA-based & 2018 & Video & $300$ real and $300$ fake videos. & \cite{guera2018deepfake} \\ \hline

\cite{khodabakhsh2018} & Fake Face in the Wild & 2018 & Video & $150$ fake videos. & \cite{khodabakhsh2018} \\ \hline

\cite{rossler2018faceforensics} & FaceForensics & 2018 & Video & $2008$ fake videos. & \cite{zhang2021detecting} \\ \hline

\cite{rossler2019faceforensics} & FaceForensics++ & 2018 & Video & $1,000$ real and $4,000$ fake videos. & \cite{moritz22} \cite{Shobha23} \cite{sun23} \cite{wang2023deep} \\ \hline

\cite{li2018ictu} & UADFV & 2018 & Video & $49$ real and $49$ fake videos. & \cite{wang2021novel} \\ \hline

\cite{korshunov2019vulnerability} & Faceswap-GAN & 2019 & Video & $640$ fake videos & \cite{korshunov2019vulnerability} \\ \hline

\cite{korshunov2018deepfakes} & Deepfake-TIMIT & 2019 & Video & $320$ real and $640$ fake videos. & \cite{zi2020wilddeepfake} \cite{umur20} \\ \hline

\cite{dolhansky2019deepfake} & DFDC-preview & 2019 & Video & $5,244$ fake videos. & \cite{das2023comparative} \cite{wang2023deep} \\ \hline

\cite{realAndFake:2019} & Real and Fake Face Detection & 2019 & Image & $1,081$ real and $960$ fake images. & \cite{Qurat:21} \cite{rafique2021deepfake} \\ \hline

\cite{li2020celeb} & Celeb-DF & 2020 & Video & $590$ real and $5,639$ fake videos. & \cite{Shobha23} \cite{raza2023holisticdfd} \cite{heo23} \cite{lin2023deepfake} \cite{Khalid:2023} \cite{wang2023deep} \\  \hline

\cite{zi2020wilddeepfake} & WildDeepfake & 2020 & Video & $3,805$ real and $3,509$ fake videos. & \cite{lin2023deepfake} \\ \hline
\end{tabular}}
\end{table}

\section{Discussion and Open Issues}
\label{s.oi}

Recent advances in fake content generation methods have gained an ever-growing concern from several legislative and regulatory authorities because of the ill use of counterfeit multimedia for illegal and public opinion manipulation. Seeking and categorizing state-of-the-art methodologies for deepfake detection goals is critical to identify the most appropriate methods to predict the actions toward a dangerous political and social instability scenario. In this sense, significant research has been reported to review and examine the well-established image and video manipulation approaches, namely those that rely on deep learning-based methods, as stated in Section~\ref{s.introduction}. 

\subsection{Recent Architectures Overview}
\label{ss.rao}

We analyzed several novel studies concerning deepfake detection using the most recent deep learning-inspired architectures, providing a more detailed start-of-the-art review.

Recent studies usually include several transfer learning-based approaches to prevent the computational load of retraining the deep learning models in massive amounts of deepfake images and videos. In this context, Sengur et al.~\cite{sengur2018} attempted to capture the generalization of pre-trained AlexNet and VGG16 models without retraining them on new images of manipulated faces. Despite the encouraging results presented in the study, fine-tuning the model's parameters is often necessary to capture intrinsic aspects of new images to improve the model's robustness on new collected features, thus increasing the ability to recognize specific forged elements. Therefore, a more comprehensive analysis is required to compare the model's effectiveness with and without fine-tuning procedure.


Residual features are essential to improve deepfake detection. El Rai et al.~\cite{elrai2020} proposed a novel approach to capture the potential noise disturbance from any video manipulation procedure. Despite being a simple approach, the main drawback regards using a small number of videos for training and evaluating the CNN model. Moreover, since individual frames appear to be considered independently as input to the model, the final decision regarding the whole video's authenticity remains unclear. Similarly, the work presented by Mo et al.~\cite{mo2018fakeface} also comprised the residual features obtained from a single high pass filter applied to the images, thus differing from the former that computes the residual noise as the difference from the original and the corresponding smoothed images. Regardless, temporal analysis is still necessary for a broader analysis of the fake aspects' interdependence across the sequences of video frames. As such, recurrent models appeared to gather the temporal data and transform them into a collection of time-series information to capture the forged sequence intercorrelation. In this context, Wang and Dantcheva~\cite{wang3dcnn2021} reported promising results considering the temporal analysis of the entire video. The authors emphasized the importance of 3D CNN models for capturing features that rely on the whole video's motion sequence, thus increasing the ability to identify the evidence of any manipulation on specific frames. Despite the low performance on test videos from a different manipulation technique, the reported results confirmed the superiority of the 3D-CNN against the baselines used for comparison. Furthermore, recent studies reported outstanding results in temporal learning domain~\cite{saif2022timedist,elpeltagy2023novel,sun23}.

On the other hand, Wesselkamp et al.~\cite{vera22} provide evidence for developing robust and non-easily deceived models to detect deepfake images more effectively. The authors presented a new class of simple attacks to evade deepfake detection by removing GAN artifacts from the frequency spectrum of the images. The generative network may use one of the selected attacks to avoid detection depending on the combination of the dataset, GAN, and detection method. The authors showed a simple but effective procedure based on the image frequency domain to mislead the deepfake detection. It shows the main concerns regarding the system's security toward an effective approach which may be deceived a priori using a new class of manipulative strategies aiming to refine the forgery of the faces and remove the evidence of possible manipulation. Therefore, it is an urge for novel research which may counteract deceived attacks and seek new features of face manipulation using the frequency domain.

Vision Transformer (ViT) has shown a potential design for several image classification tasks. Based on its counterpart Transformer model initially proposed for Natural Language Processing tasks, the inherent sequential analysis used in the ViT architecture provides the local and global extraction of the features by combining the image patches in sequential order arrangement. An attempt to incorporate the ViT architecture into the deepfake detection context was presented by Wodajo and Atnafu~\cite{wodajo2021}, which used the feature maps provided by the convolutional layer of the VGG CNN architecture. By doing so, the vision transformer is not limited to the patch sizes described in the original work of Dosovitski~\cite{dosovitskiy2020image}, thus making the model more flexible for other CNN architectures with different output feature map sizes.

In social engineering, deepfake algorithms have raised new chances to earn unauthorized access to private and confidential information, leading to a pressing concern considering credentials manipulation. Despite being a recent form of attack, deepfake phishing proved to be a dangerous threat as a criminal instrument for obtaining financial benefits~\cite{loeffler2019}. In biometric systems, deepfake may pose a significant threat to access control through spoofing of face biometrics, as pointed out by Wojewidka~\cite{wojewidka2020}. Although irrelevant to deepfake applications, fake content detection is critical to prevent several forms of credential tampering. Popular biometric systems often rely on fingerprint analysis, face identification, and iris features in modern applications. The work of Goel et al.~\cite{goel2020} showed promising results for counterfeit fingerprint detection using a single deep-learning architecture. Despite the possible broadening investigation of spoofing techniques to other applications rather than the face point-of-view, we focused only on face forgery detection for their inherent application in the deepfake context.

\subsection{Opportunities and Future Challenges}
\label{ss.ofd}

Regarding the final thoughts to complement this section, we present the following opportunities and future directions for further studies, which are summarized in Figure \ref{f.challenges}:
\begin{figure}[h!]
\includegraphics[width=\textwidth]{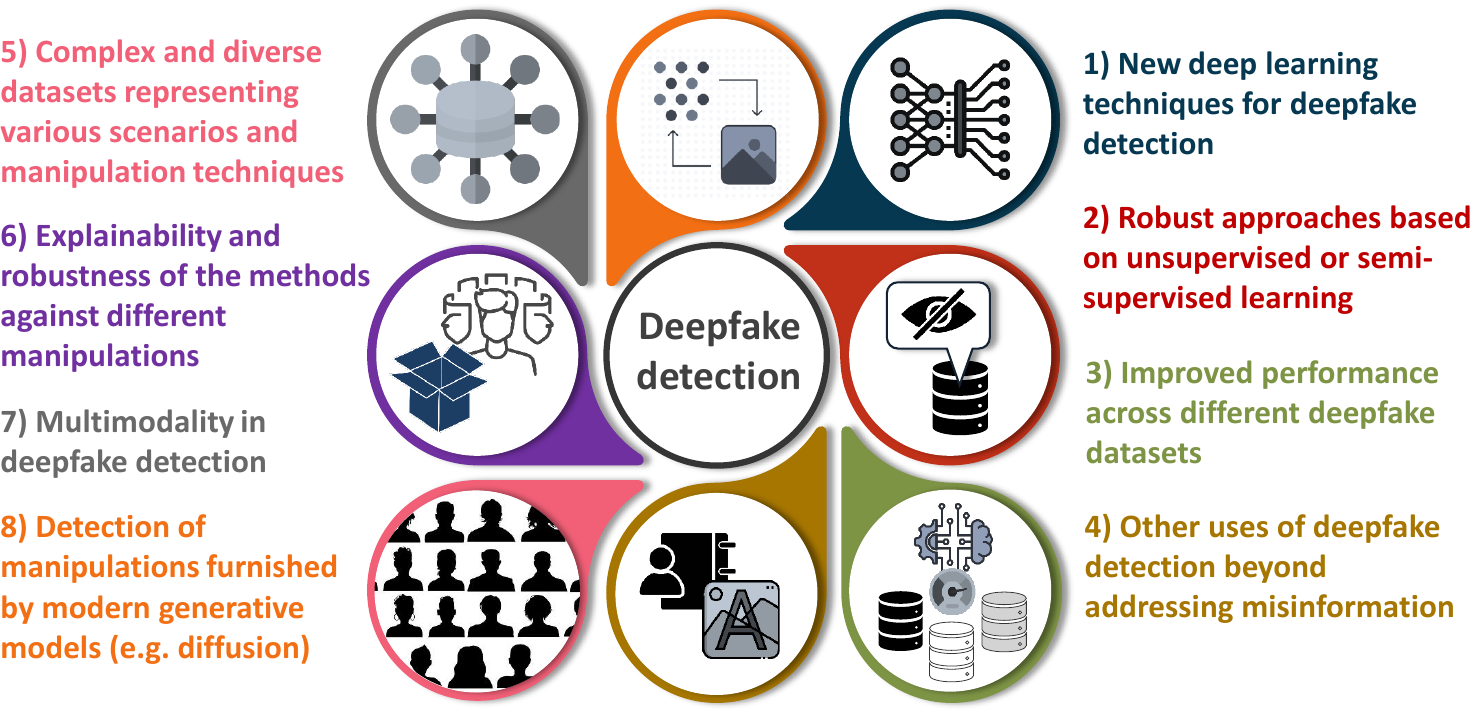}
\caption{Summary of the challenges and research directions identified in this work.}
\label{f.challenges}
\end{figure}

\begin{enumerate}[leftmargin=*]
	\item Complexity and realism of deepfake methods are actively refined as a consequence of the advances in deep learning techniques. Therefore, it is highly desirable to follow the trends toward the analysis of large amounts of data and explore dynamic approaches to identify and extract additional features from recent sources of information.
	\item Deepfake generation is a constant process and continuously evolves due to social networks and the rapid spreading of new information. Therefore, more robust approaches based on unsupervised or semi-supervised learning are particularly essential to avoid the time-consuming and laborious manual annotation of massive amounts of new data.
	\item Regarding the cross-dataset and ablation experiments, most of the examined studies reported decreased performance when the models were trained and validated on different deepfake datasets. Moreover, there is a challenge in reaching the same best-performing accuracy for different fake production methods. Most related studies provided lower accuracy for the NeuralTextures and FaceShifter manipulation of the FaceForensics++ dataset~\cite{Ilyas2022,saif2022timedist,sun23}. It shows an urge to explore more complex forgery traits produced by several manipulation techniques and the challenges towards developing more accurate methods for fake face detection.
	\item In addition, we can leverage knowledge in deepfake detection for other uses rather than just addressing misinformation. For instance, we can harness the power of generative artificial intelligence for authorship attribution of deepfake content~\cite{uchendu2023toproberta} and watermarking to embed a signature in an image and prevent it from deepfake tampering~\cite{Zhao2023watermarking,Luochen2021watermarking}. This capacity is remarkably essential in the context of generative modeling and AI-based content.
	
	\item The predominant use of English, Japanese, and Chinese datasets emphasizes the need to incorporate a wider diversity of languages and cultural contexts. Further, current datasets have significant limitations regarding quality and variety, demanding the development of more complex and diverse datasets capable of representing various scenarios and manipulation techniques. It is observed that most studies concentrate on the detection techniques themselves, neglecting the importance of the datasets to train and evaluate such techniques.
	
	\item Similarly, it is necessary to develop detection systems capable of generalizing to different techniques, ensuring that such methods are robust enough to handle a variety of manipulation techniques and generation models. At the same time, the lack of explainable artificial intelligence techniques stands out in most studies, indicating the relevance of transparency and understandability to users. Such transparency proves crucial to building trust in detection systems and mitigating the spread of misinformation.

	\item There is a trend to combine visual and audio information to improve the accuracy of detecting forgery faces in videos. However, only a few studies have been explored to reveal the benefits of multi-modal approaches~\cite{hao2022deepfake,elpeltagy2023novel}. Therefore, more research is encouraged to assess the impacts of using audio characteristics on deepfake detection performance.
	\item Finally, recent advances in generating realistic images produced by diffusion models show promise even when compared to images generated by GANs. In a brief explanation, diffusion models~\cite{Rombach:2022, Ho:2020, Sohl-Dickstein:15} are governed by two processes. The forward process is described by a Markov chain in which Gaussian noise is gradually added to a given image at each iteration. Moreover, the backward process involves learning the denoising process, i.e., from a noise-corrupted image to a clear image. Currently, there are still very few studies carried out in the sense of using diffusion models to generate deepfake~\cite{Mandelli:2022, Jeong:2022} and, in the same sense, few studies involving the detection of deepfake also generated by diffusion models.

\end{enumerate}

\section{Conclusions}
\label{s.cn}

This work highlights the most significant research in the last years regarding deepfake detection through deep learning techniques. Besides, it also presents the most relevant advances in the field and the main challenges and future trends. A brief analysis of the works presented in this study may deduce a correlation between fake news subjects and deepfake content, for deepfake production is intrinsically related to video manipulation, which denotes a complement to fake news content. Hence, it is necessary to investigate several studies that simultaneously merge fake news and deepfake.

An alarming concern regarding deepfake production regards the fast development of generative networks, implying more realistic and high-quality images. Such a tendency infers an increasing challenge, making detection even more difficult. The emergence of new intelligent algorithms and the evolution of existing ones may be the most plausible inclination to tackle the problem.

As stated in the last section, the field's future direction comprises more robust and dynamic approaches to deal with realistic deepfake content. Such models should identify and extract new features and analyze large amounts of data for better detection rates. Moreover, semi-supervised learning techniques should help to deal with the fast evolution of deepfake generators and the spread of their content in social networks. In this sense, future studies demand a more dynamic procedure or even the combination of supervised and unsupervised learning to rapidly identifying and actively tracking the patterns related to the modern and complex fake content production algorithms without requiring large amounts of data.

\section*{Acknowledgments}
\begin{sloppypar}
The authors are grateful to Conselho Nacional de Desenvolvimento Cient\'{i}fico e Tecnol\'{o}gico (CNPq), Brazil grants \#429003/2018-8, \#307066/2017-7 and \#427968/2018-6, Funda\c{c}\~ao de Amparo \`{a} Pesquisa do Estado de S\~ao Paulo (FAPESP), Brazil grants  \#2021/05516-1, \#2013/07375-0, \#2014/12236-1, \#2023/10823-6, and \#2019/07665-4, Petrobr\'{a}s, Brazil grant \#2017/00285-6, and 06/2023 PROPe-UNESP for their financial support. J. Del Ser acknowledges funding support from the Basque Government through ELKARTEK and EMAITEK funds as well as the Consolidated Research Group MATHMODE (IT1456–22).
\end{sloppypar}

\bibliography{references}

\pagebreak
\noindent \textbf{Leandro Aparecido Passos} is graduated in Informatics, has M.Sc. and Ph.D. in Computer Science, and worked as a post-doctorate researcher at the University of Wolverhampton (UK). Currently, is engaged as a researcher at UNESP. Has experience in Machine Learning, and most of his works employ graph- and energy-based approaches, as well as more biologically plausible algorithms.

\noindent \textbf{Danilo Samuel Jodas} has B.Sc. and M.Sc. in Computer Science, and holds a Ph.D. in Informatics Engineering from the Faculty of Engineering of the University of Porto (FEUP), Portugal. He has experience in image processing and machine learning. He is currently a postdoctoral researcher at the S\~{a}o Paulo State University.

\noindent \textbf{Kelton Augusto Pontara da Costa} is graduated in Systems Analysis, has M.Sc. and Ph.D. in Computer Science, and Post-doctoral in Computer Networks by UNICAMP and UNESP. Currently works as Reseacher/Professor at UNESP and as M.Sc and PhD advisor at UNESP. Has experience in Computer Science with emphasis in Cybersecurity and is a senior member of the IEEE.

\noindent \textbf{Luis Antonio de Souza J\'{u}nior} has B.Sc. and M.Sc. in Computer Science by UNESP. Current Ph.D. student by the "Universidade Federal de São Carlos (UFScar)", Member of the Recogna group (UNESP-Bauru) and fellow member of ReMIC group from the Technical University of Applied Sciences - Regensburg, Germany. 

\noindent \textbf{Douglas Rodrigues} majored in Business Management and Informatics at FATEC - Botucatu's Faculty of Technology, SP, Brazil~(2009). In 2014, he received his M.Sc. in Computer Science from S\~ao Paulo State University (UNESP). In 2019, he received his Ph.D. in Computer Science from the Federal University of S\~ao Carlos (UFSCar), SP, Brazil. Currently, he is working as post-doctorate researcher at S\~ao Paulo State University (UNESP), SP, Brazil. His interests include machine learning, single and multi-objective optimization.

\noindent \textbf{Javier Del Ser} defended his first doctoral thesis (Cum Laude) in Control Engineering and Industrial Electronics at the University of Navarra (2006), and a second doctoral thesis in Information and Communication Technologies (also Cum Laude and awarded the Extraordinary PhD Award) at the University of Alcala de Henares (2013). He is a Research Professor and Chief Scientist in Artificial Intelligence at TECNALIA RESEARCH \& INNOVATION, and an Adjunct Professor at the Department of Communications Engineering of the University of the Basque Country (EHU/UPV). He is also a distinguished Professor at the University of Granada (Spain). His research interests are in Artificial Intelligence, machine learning and deep learning applied to practical modeling and optimization tasks for problems arising from different sectors, including industry, health, telecommunications, transportation, energy, and mobility, among others. He has authored more than 450 scientific contributions to date, including 180 JCR-indexed journal articles. He has supervised 16 doctoral theses and participated in more than 50 projects and contracts. He has been listed within the top 2\% most influential AI researchers worldwide by the Stanford University and has also been part of the team that developed the R\&D strategy in Artificial Intelligence for the Government of Spain in 2019.

\noindent \textbf{David Camacho} is currently a Full Professor at the Computer Systems Engineering Department of the Technical University of Madrid (Spain), and the Head of the Applied Intelligence \& Data Analysis group. He received a Ph.D. with honors in Computer Science from Universidad Carlos III de Madrid in 2001. He has published more than 350 journals, books, and conference papers. His expertise comprises: Big Data; Machine Learning: Clustering, Hidden Markov Models, Classification and Deep Learning; Computational Intelligence: Evolutionary computation, Swarm Intelligence; Pattern and Process modeling and mining; Graph Computing and Social Mining, and Data Analysis for complex industrial applications for companies, such as: Airbus Defence \& Space, Codice Technologies, ImpactWare, or Jobssy S.L among others.

\noindent \textbf{Jo\~{a}o Paulo Papa} received his B.Sc. in Information Systems and his M.Sc. and Ph.D. in Computer Science. Had worked as a post-doctorate at UNICAMP and as a visiting scholar at Harvard. Actually, is a Computer Science Professor at UNESP. Also, the recipient of the Alexander von Humboldt research fellowship and is a senior member of the IEEE.

\end{document}